\documentclass[runningheads,a4paper]{llncs}

\usepackage{amsmath,amsfonts,bm}

\def\eqref#1{equation~\ref{#1}}

\def\1{\bm{1}}

\DeclareMathAlphabet{\mathsfit}{\encodingdefault}{\sfdefault}{m}{sl}
\SetMathAlphabet{\mathsfit}{bold}{\encodingdefault}{\sfdefault}{bx}{n}

\usepackage[T1]{fontenc}
\usepackage[utf8]{inputenc}
\usepackage{graphicx}
\usepackage{booktabs}
\usepackage{amssymb}
\usepackage{nicefrac}
\usepackage{microtype}
\usepackage{xcolor}
\usepackage{longtable}
\usepackage{float}
\usepackage{subcaption}
\usepackage{tcolorbox}
\usepackage{listings}
\usepackage{enumitem}
\usepackage{wrapfig}
\usepackage{tabularx}
\usepackage{bbm}
\usepackage{multirow}
\usepackage{colortbl}
\usepackage{makecell}
\usepackage{tikz}
\usepackage[linesnumbered,ruled,vlined]{algorithm2e}
\usepackage{textgreek}
\usepackage[hidelinks]{hyperref}

\newcommand{\citep}[1]{\cite{#1}}

\newcommand{\system}{{\sc Scorpio}}
\newcommand{\myline}[1]{{\medskip\noindent\textbf{#1}}}

\begin{document}

\title{\textsc{Scorpio}: Serving Right Requests at the Right Time for Heterogeneous SLOs in LLM Inference}
\titlerunning{Scorpio: Serving Right Requests at the Right Time}

\author{Yinghao Tang\inst{1} \and
Tingfeng Lan\inst{2} \and
Bo Pan\inst{1} \and
Xiuqi Huang\inst{1}\textsuperscript{*} \and
Hui Lu\inst{3}\textsuperscript{*} \and
Wei Chen\inst{1}}
\authorrunning{Y. Tang et al.}
\institute{State Key Lab of CAD\&CG, Zhejiang University
\and Department of Computer Science, University of Virginia
\and Department of Computer Science and Engineering, The University of Texas at Arlington\\
\textsuperscript{*}Corresponding authors: \email{huangxiuqi@zju.edu.cn (primary), hui.lu@uta.edu}}

\maketitle

\begin{abstract}
   Large Language Model (LLM) serving increasingly underpins online Web services such as conversational agents, Web search, and programming assistants, where requests carry heterogeneous Service Level Objectives (SLOs) such as Time to First Token (TTFT) and Time Per Output Token (TPOT). Existing LLM serving systems prioritize maximum throughput and treat all requests uniformly, which leads to suboptimal SLO attainment. This paper introduces \system{}, an SLO-oriented LLM serving system designed to maximize system goodput and SLO attainment for workloads with heterogeneous SLOs. Our core insight is to exploit SLO heterogeneity for adaptive scheduling across admission control, queue management, and batch selection. \system{} features a TTFT Guard, which employs least-deadline-first reordering and rejects unattainable requests, and a TPOT Guard, which utilizes a VBS-based admission control and a novel credit-based batching mechanism. Both guards are supported by a predictive module. Evaluations demonstrate that \system{} improves system goodput by up to 14.4$\times$ and SLO adherence by up to 46.5\% under high load compared to state-of-the-art baselines.

\keywords{LLM serving \and Model-as-a-Service \and Service Level Objectives \and Request scheduling \and Web service quality.}
\end{abstract}

\section{Introduction} \label{sec:intro}

Large Language Models (LLMs) have become core infrastructure for Web-delivered AI services, served to end users over the Web through Model-as-a-Service (MaaS) platforms that power conversational agents \citep{manus}, Web search and deep-search engines \citep{grok3}, and programming assistants \citep{copilot}. As these services scale to millions of concurrent Web requests, the inference backend behind each HTTP endpoint becomes the central performance bottleneck for user-facing quality. To handle the substantial computational requirements of LLM inference, state-of-the-art serving systems such as vLLM \citep{vllm} and SGLang \citep{sglang} employ advanced techniques, including continuous batching \citep{Orca}, paged attention \citep{vllm}, chunked prefilling \citep{chunkedPrefill}, etc. These optimizations markedly improve inference throughput and resource utilization.

Despite these efficiency improvements, existing LLM serving systems \citep{vllm,Orca,sglang,tensorrt-llm} predominantly prioritize maximum throughput, often neglecting Service Level Objectives (SLOs), such as time-to-first-token (TTFT) and time-per-output-token (TPOT). Typically, these systems greedily admit and serve incoming requests \citep{miao2023efficientgenerativelargelanguage}, without deep visibility into the SLO requirements of requests. Simultaneously, SLO requirements across different applications are inherently heterogeneous \citep{QueueManagement}. For instance, programming assistants often demand low latency for real-time response, whereas chatbots tolerate higher latency provided the generation rate exceeds human reading speed \citep{Adaserve}. However, existing LLM serving systems treat all requests equally in all scheduling stages. This undifferentiated handling leads to suboptimal SLO attainment. These problems necessitate a fine-grained and adaptive scheduling method to support heterogeneous SLOs.

In this paper, we focus on designing an SLO-oriented LLM serving system that is tailored for heterogeneous SLOs, with the goal of maximizing both system goodput and SLO attainment. \textbf{Our key insight is that SLO heterogeneity is not a constraint to tolerate but a resource to exploit: by scheduling the right request at the right time across queue management, admission control, and batch selection, the system converts slack in loose-SLO requests into headroom for tight-SLO requests, lifting system-level SLO attainment without extra hardware.} From the TTFT perspective, requests with looser SLO can be served later. From the TPOT perspective, requests with looser TPOT can skip some iterations of generation, leaving more resources for requests with tighter TPOT.

Based on this insight, we design TPOT Guard and TTFT Guard to handle the heterogeneous TPOT and TTFT SLOs, respectively. For the TPOT Guard, we first define a core concept of TPOT-relative Proportionality (TRP), which quantifies the heterogeneity of TPOT SLOs. Utilizing TRP, we propose a novel VBS-based Admission Control mechanism and a Credit-based Batching mechanism. The former is to control the admission of requests to prevent mass TPOT violations due to overwhelming request ingress. The latter mechanism adaptively batches requests according to their per-request TPOT SLO, achieving system-level high TPOT SLO attainment. For the TTFT Guard, we implement a simple but effective least-deadline-first reordering strategy that prioritizes requests nearing their TTFT deadlines. Additionally, under heavy load, we reject the requests that are unattainable for their TTFT SLOs. To provide decision-making support for these two modules, we develop a predictor module consisting of a sequence length predictor and two analytical models. By orchestrating these complementary mechanisms in concert, \system{}  enables robust support for diverse workloads.

Our contributions are summarized as follows:
\begin{itemize}[leftmargin=10pt, itemsep=0pt]

  \item \textbf{New setting.} We co-optimize \emph{heterogeneous} TTFT and TPOT SLOs within a single serving system, a setting that underpins multi-tenant Web services such as chatbots, coding assistants, and search, where different applications demand sharply different latency guarantees.

\item \textbf{TPOT Guard.} We design a TPOT Guard built on a novel TPOT-relative Proportionality (TRP) abstraction, pairing Virtual-Batch-Size admission control with credit-based batching to drive per-request TPOT adherence under load.

\item \textbf{TTFT Guard.} We design a TTFT Guard combining least-deadline-first reordering with proactive rejection of unattainable requests, preventing starvation while honoring heterogeneous TTFT deadlines.

\item \textbf{Predictive module.} We build a lightweight predictive module, namely a fine-grained sequence-length predictor and two analytical latency models that estimate TPOT and TTFT with R$^2$ above 0.9, at under 1\% scheduling overhead.

\item \textbf{System and results.} We implement \system{} on vLLM and evaluate it on Llama-3.1-8B and Gemma-2-27B across ShareGPT, LMSYS-Chat-1M, and real Azure production traces. \system{} improves system goodput by up to \textbf{14.4$\times$} and SLO adherence by up to \textbf{46.5\%} over state-of-the-art baselines, with negligible (under 1\%) scheduling overhead.

\end{itemize}

\begin{figure}[t]
  \centering
  \includegraphics[width=0.82\textwidth]{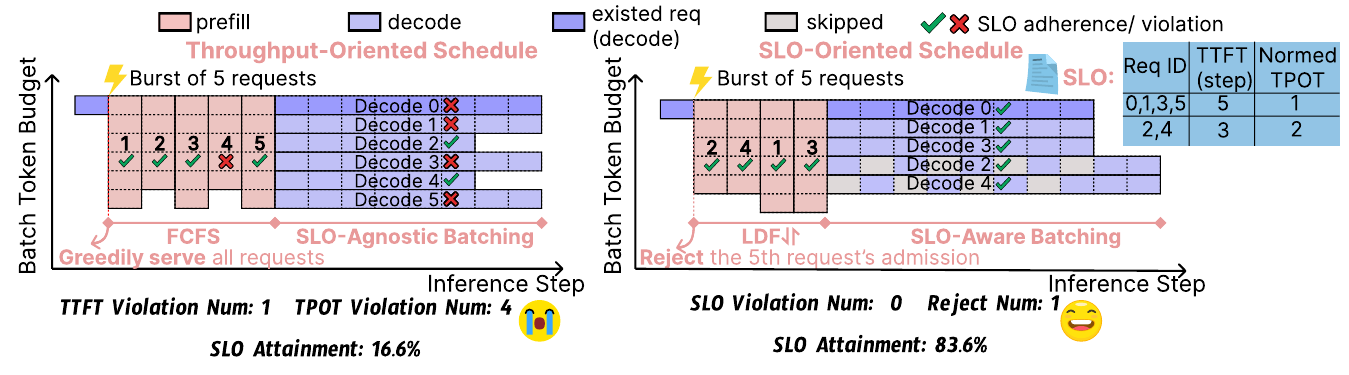}
  \caption{Comparison of throughput-oriented and SLO-oriented scheduling approaches. The throughput-oriented scheduler greedily admits and processes requests without considering per-request SLOs. During the prefill phase, request 4 violates its SLO constraint (4 steps> 3 steps). The SLO-oriented scheduler prevents such violations through least-deadline-first prioritization. During the decode phase, we assume a normalized decode step time as 0.25$\times$ BatchSize. The throughput-oriented scheduler batches all requests in each step (BatchSize=6), causing requests 0, 1, 3, and 5 to violate their TPOT constraints (each step consumes a time of 1.5). In contrast, the SLO-oriented scheduler rejects unattainable requests (e.g., request 5) and implements an adaptive fine-grained batching strategy (BatchSize=4). This strategy allows requests with looser TPOT SLOs (requests 2 and 4) to skip certain iterations, ensuring all admitted requests satisfy their TPOT constraints.}
  \label{fig:example}
\end{figure}

\section{Background and Problem Formulation} \label{sec:bac}

A typical LLM generative inference task has two stages: i) the prefill stage, which takes a prompt sequence to generate the first output token; and ii) the decoding stage, which generates new tokens autoregressively. The quality of LLM service is typically evaluated by two key metrics: \textit{time to first token} (TTFT) and \textit{time per output token} (TPOT) \citep{pddisaggreate}. TTFT captures latency for generating the first output token after a request is received, while TPOT sets an upper bound on the average latency for generating subsequent tokens. Meeting SLOs plays a key role in providing high-quality LLM services and has been fully researched in other fields like cloud and edge computing \citep{slo1,slo2}. However, this problem has not been well explored in LLM serving. Also, from the perspective of Model as a Service (MaaS) providers, different applications and users have different SLO requirements \citep{QueueManagement,sloserve}. In a Web setting, these per-request SLOs are the latency contracts a MaaS endpoint exposes to downstream Web applications, so meeting them heterogeneously lets a single shared backend honor the diverse quality-of-service expectations of distinct Web tenants \citep{jiang2025hierarchicalpredictionbasedmanagementlmaas}.  This inspires us to explore the heterogeneous SLO attainment problem in LLM serving.

We define the heterogeneous SLO attainment problem as follows:  During a time interval $T$, there is a sequence of user requests $\mathcal{R} = \{r_1, r_2, \ldots, r_N\}$ arriving. An LLM inference system processes the requests with a scheduling policy $\pi$. A request $r_i \in \mathcal{R}$ with its TTFT SLO threshold $S_{TT}(r_i)$ and TPOT SLO threshold $S_{TP}(r_i)$ is considered SLO-compliant if it satisfies both $TTFT(r_i) \le S_{TT}(r_i)$ and $TPOT(r_i) \le S_{TP}(r_i)$. Let  $\mathcal{R}^{good}(\pi) \subseteq \mathcal{R}$ be the subset of those requests that are SLO-compliant. Note that the denominator $|\mathcal{R}|$ counts all arriving requests, so rejected requests are counted as SLO-unmet and never inflate either metric. Then the system goodput and SLO adherence rate can be defined as:
\begin{gather}
    \text{Goodput}(\pi) = \frac{|\mathcal{R}^{good}(\pi)|}{T} \\ %
    \text{Adherence}(\pi) = \frac{|\mathcal{R}^{good}(\pi)|}{|\mathcal{R}|} \quad 
\end{gather}

The objective is to design an online scheduling policy $\pi$ aiming to achieve:
\begin{equation} \label{eq:multi_objective_optimization}
    \max_{\pi} \quad \left( \lim_{T \to \infty} \mathbb{E}[\text{Goodput}(\pi, T)], \quad \lim_{T \to \infty} \mathbb{E}[\text{Adherence}(\pi, T)] \right)
\end{equation}

\begin{figure}[t]
  \centering
  \includegraphics[scale=0.085]{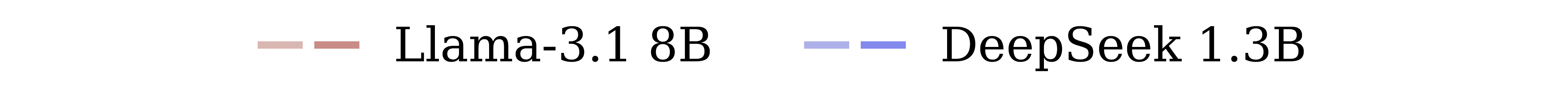} \\
  \begin{subfigure}[b]{0.32\textwidth}
    \centering
    \includegraphics[scale=0.085]{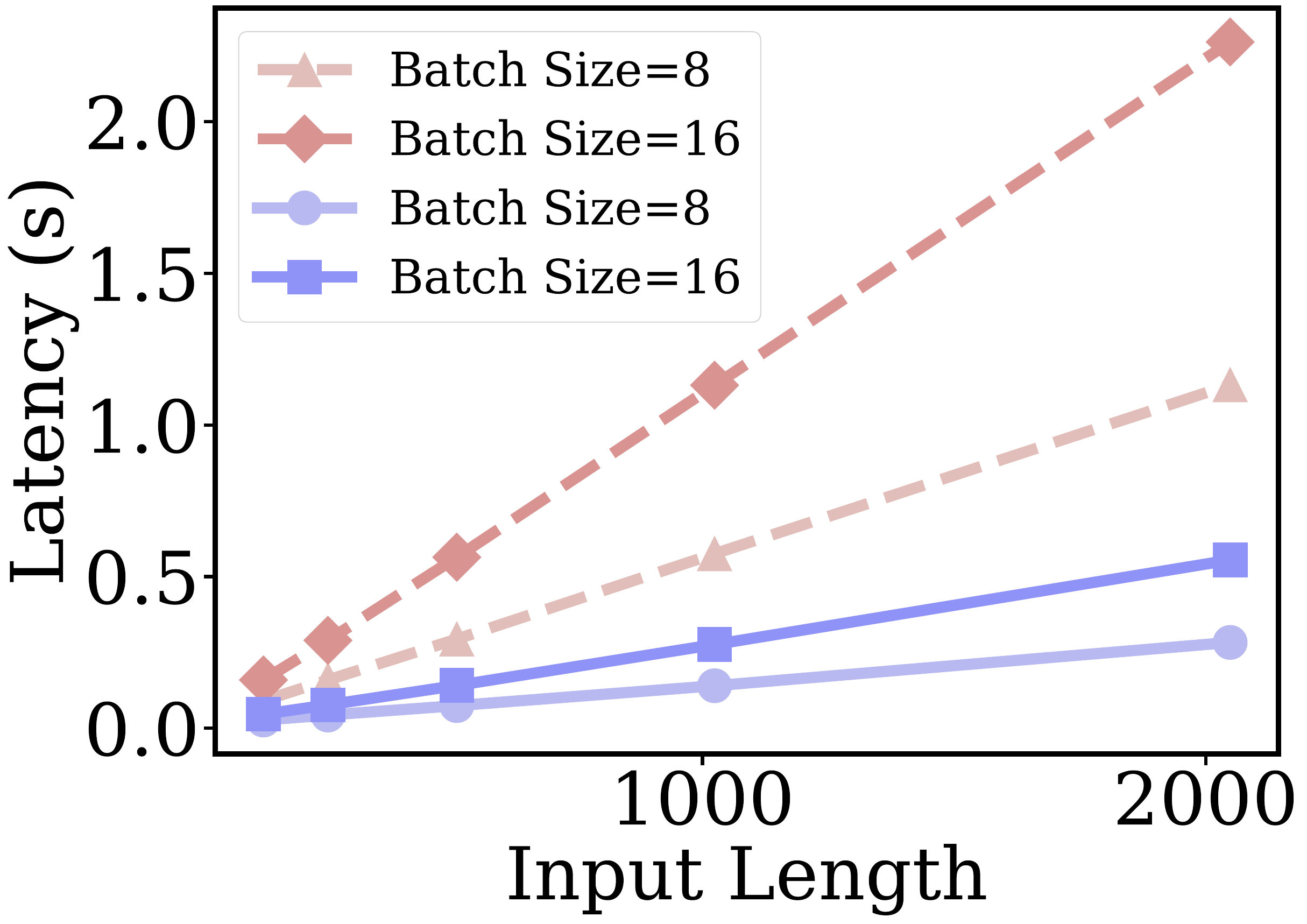}
    \caption{\small{Prefill Latency vs Input Length}}
    \label{subfig:prefill_latency_vs_input_length}
  \end{subfigure}
  \hfill
  \begin{subfigure}[b]{0.32\textwidth}
    \centering
    \includegraphics[scale=0.085]{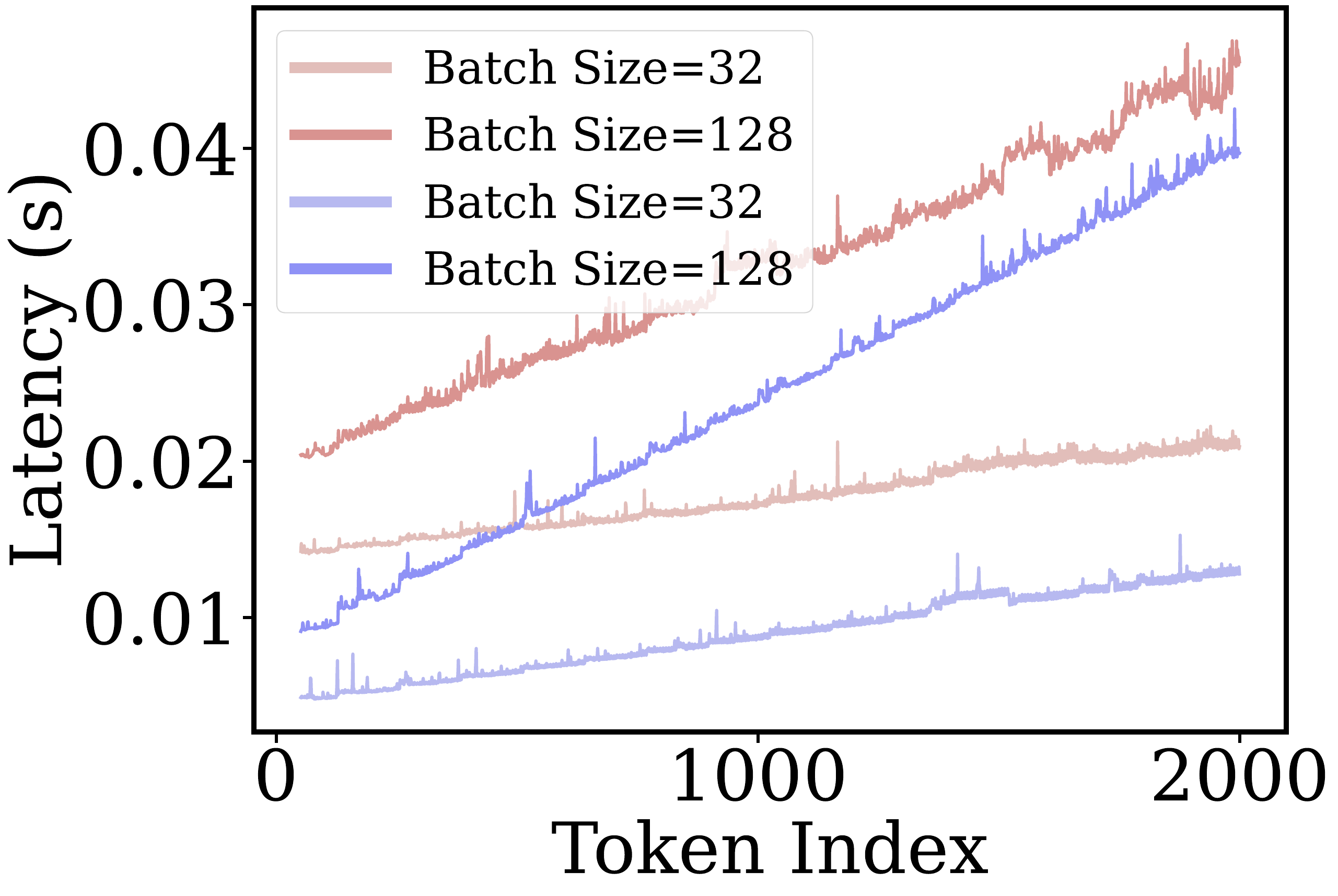}
    \caption{\small{Decode Latency vs Token Index}}
  \end{subfigure}
  \hfill
  \begin{subfigure}[b]{0.32\textwidth}
    \centering
    \includegraphics[scale=0.085]{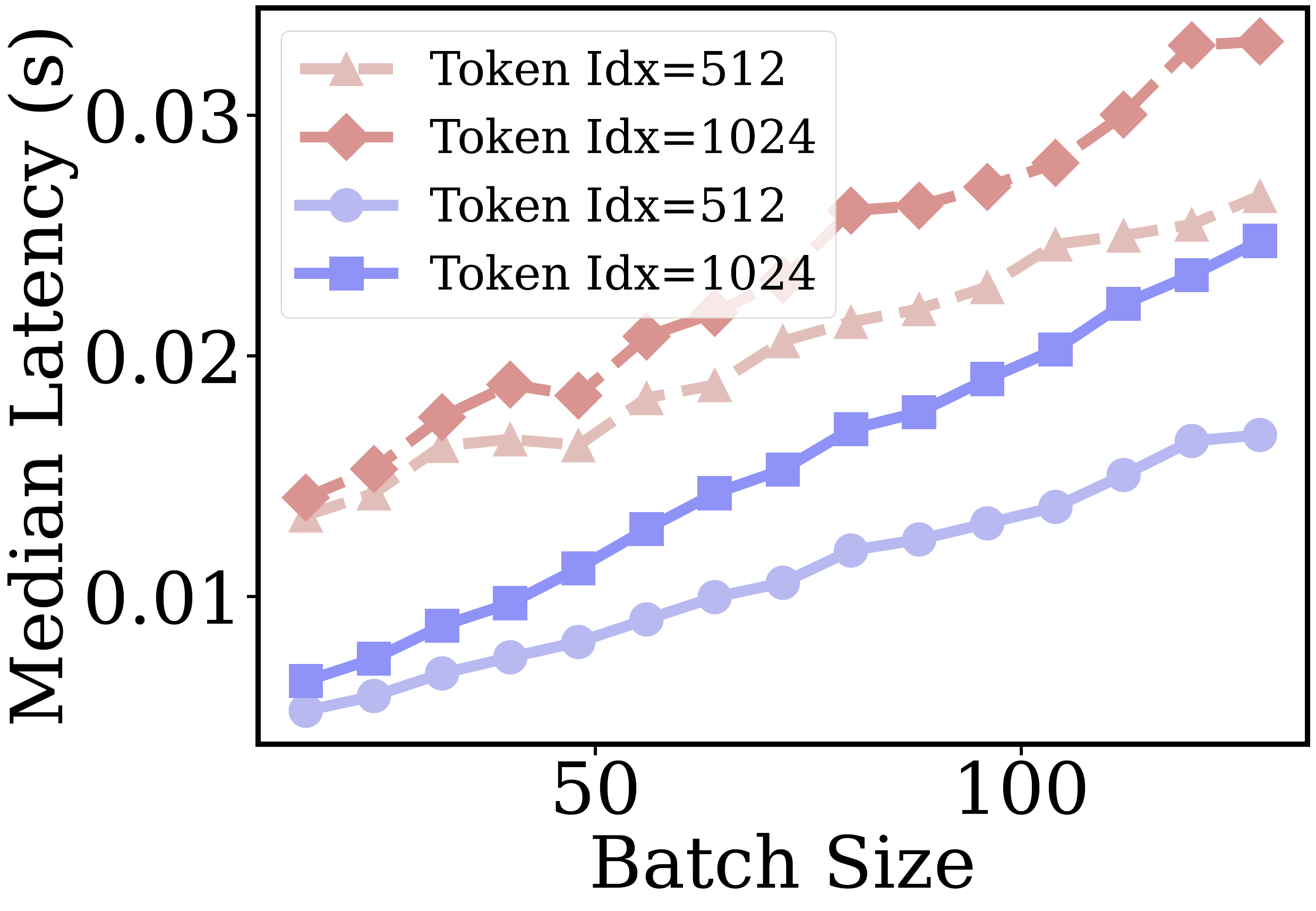}
    \caption{\small{Decode Latency vs Batch Size}}
  \end{subfigure}
  \caption{Impact of input sequence length (a), batch size (b), and output token index (c) on prefill latency and inter-token latency (ITL).}
  \label{fig:observation}
\end{figure}
\section{Method} \label{sec:method}

\subsection{System Overview} \label{subsec:system_overview}

\system{} is a system-algorithm co-design framework designed to maximize goodput and SLO adherence. \system{} consists of three key components: 1) a predictor, 2) a TTFT Guard, and 3) a TPOT Guard. The two guards leverage predicted SLO metrics to explicitly handle heterogeneous request requirements.

\begin{figure}[!t]
    \centering
    \includegraphics[width=0.9\linewidth, trim= 0 0 0 0, clip]{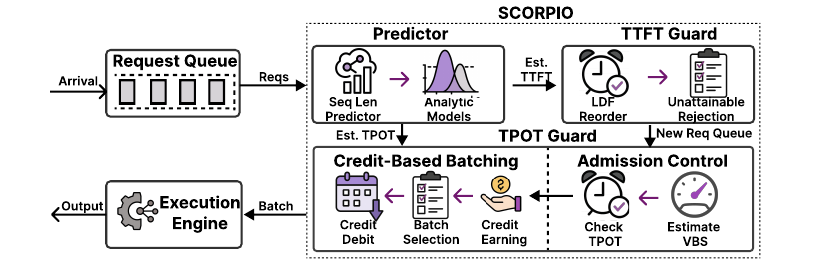}
    \caption{The workflow of \system{}.}
    \label{fig:overview}
\end{figure}

Figure \ref{fig:overview} shows the workflow of \system{}: when a request arrives, the predictor (\S \ref{subsec:predictor}) first predicts the output sequence length, which is used by two analytic models to estimate the TPOT and TTFT.  Given the estimated information, the TTFT and TPOT Guards make scheduling decisions accordingly. Specifically, the TTFT Guard (\S \ref{subsec:heterogeneous_ttft_gurantee}) first reorders the requests according to their TTFT SLO deadline (i.e., Least Deadline First Reordering ). Additionally, it uses the estimated TTFT to reject requests that are unattainable with respect to their TTFT SLOs. Given the new reordered requests, the TPOT Guard (\S \ref{subsec:heterogeneous_tpot_gurantee}) uses the estimated TPOT to decide whether to admit the requests into the running batch (i.e., VBS-based Admission Control Mechanism ). Then, it employs a Credit-based Batching Mechanism to select which requests to batch in each processing iteration. Lastly, the selected requests are batched and processed in the execution engine.

\subsection{Predictor} \label{subsec:predictor}


\myline{Sequence Length Predictor.} We design a lightweight predictor for output sequence length with minimal overhead. In contrast to prior work \citep{s3,bertprediction}, which typically uses coarse-grained bucketing (e.g., 10 bins), we conduct a systematic analysis of bucketing strategies and show that medium-grained binning (e.g., 100 bins) provides superior accuracy: for Llama3-8B on ShareGPT, 100 equal-width bins give Kendall's $\tau$ of 0.54 versus 0.25 at 10 bins. This design improves on prior approaches along two axes: (i) it corrects the class imbalance that inflates overall accuracy while degrading prediction on minority classes, and (ii) it sustains resolution as maximum context lengths grow in modern LLMs. We fine-tune OPT-125M \citep{optseries} as a length-bin classifier on 20K ShareGPT and LMSYS-Chat-1M samples and co-locate it with the serving engine.

\myline{SLO Metric Estimation.} We observe two key statistical correlations from our empirical analysis (Figure \ref{fig:observation}, profiled on a representative model as these trends are consistent across models): (1) inter-token latency (ITL) is positively correlated with batch size and average sequence length, and (2) prefill time is positively correlated with prompt length. Based on these observations, we develop analytical models to estimate TPOT and TTFT for scheduling decisions.

For TPOT estimation, we model ITL as $\mathcal{F}(|R|, L)$ where $|R|$ is batch size and $L$ is average sequence length. Given a new request $r$ with predicted output length $P(r)$, we estimate the batch-level TPOT as the expected ITL over the next $P(r)$ generation steps. For TTFT estimation, we model prefill time as $\mathcal{G}(L_P)$ where $L_P$ is prompt length. The estimated TTFT for a waiting request is the cumulative prefill time of all preceding requests in the queue. Across both models and datasets, these estimators attain R$^2$ above 0.9 (0.90 to 0.99).

\subsection{Heterogeneous TTFT Guard} \label{subsec:heterogeneous_ttft_gurantee}

\myline{Least Deadline First (LDF) Reordering.} To handle heterogeneous TTFT SLOs, \system{} uses a simple but effective LDF reordering strategy. The deadline for a request $r_i$ is calculated as the time left to its TTFT deadline $S_{TT}(r_i)$. This strategy puts more urgent requests (with earlier TTFT deadlines) at the front of the queue, achieving better system-level TTFT attainment.

\myline{Unattainable SLO Rejection.} When the system load is excessively high, some requests will inevitably violate their TTFT SLO and need special handling. In this paper, we tag these requests as unattainable SLO requests and reject them for simplicity. These rejected requests integrate naturally with complementary cloud-side mechanisms such as cross-node migration or elastic scaling \citep{dlrover}, making \system{} a drop-in scheduling layer within a larger serving fabric.

\myline{Starvation Prevention.} The combination of LDF Reordering and Unattainable SLO Rejection inherently prevents request starvation. As time progresses, requests with approaching deadlines gain higher priority and move to the front of the queue, while requests that would inevitably miss their deadlines are proactively rejected rather than allowed to starve.

To quantitatively assess starvation prevention in our heterogeneous SLO environment, we propose a $max\_waiting\_ratio$ fairness metric to evaluate the fairness among requests. Different from absolute waiting time metrics \citep{learbyrank}, we define $max\_waiting\_ratio$ as:

\begin{equation}
    max\_waiting\_ratio = \max_{r \in R}\left(\frac{waiting\_time(r)}{S_{TT}(r)}\right)
\end{equation}

where $S_{TT}(r)$ is the TTFT SLO requirement for request $r$. This metric captures relative starvation by accounting for individual SLO, where lower values indicate better starvation prevention (\S \ref{subsec:effectiveness_analysis}).

\subsection{Heterogeneous TPOT Guard} \label{subsec:heterogeneous_tpot_gurantee}

\myline{Key Insight.} \textbf{Treating every request identically wastes the system's most valuable lever: requests with looser TPOT can be batched less frequently, returning compute to requests with tighter TPOT and converting per-request SLO slack into system-wide guarantee capacity.} As shown in Figure \ref{fig:example}, existing methods indiscriminately admit all incoming requests and process them equally in each step. This approach leads to two primary issues: 1) SLO violations for requests with strict TPOTs when contending with those having looser TPOT SLOs, and 2) when the system workload exceeds its processing capacity, greedily serving all requests causes a cascading effect where all requests fail to meet their SLOs. To address the first issue, we design a novel batching mechanism that provides differentiated batching opportunities based on a request's TPOT SLO (i.e., Credit-based Batching Mechanism). Requests with looser TPOT SLOs are offered fewer credits (opportunities) for batching, thereby skipping some execution iterations and leaving more resources for requests with tighter TPOT SLOs. For the latter issue, inspired by load control techniques in cloud computing \citep{dlrover}, we introduce a novel admission control mechanism that accounts for the heterogeneity of TPOT SLOs. On one hand, requests estimated to cause TPOT violations are rejected. On the other hand, since a request with looser TPOT SLOs is served intermittently, it can be regarded as a partial request when calculating the batch size (i.e., Virtual Batch Size) for TPOT estimation (\S \ref{subsec:predictor}). These two mechanisms together enforce TPOT SLOs in a best-effort manner, as illustrated in Algorithm \ref{alg:tpot_guarantee_mechanism}.

\myline{Credit-based Batching } offers requests of different TPOT SLOs with different credits (opportunities) for batching each iteration as mentioned above. To decide how many credits a request earns, we first introduce a concept called TPOT-relative Proportionality (TRP):

\begin{definition}[TPOT-relative Proportionality (TRP)]
    Let $S_{TP}(r)$ denote the TPOT SLO of request $r$. Given current processing iteration $t$ and the running requests set $R(t)$, the TPOT-relative Proportionality (TRP) of a request $r \in R(t)$ is defined as:
    \[ \text{TRP}(r) = \frac{\min_{r' \in R(t)} S_{TP}(r')}{S_{TP}(r)} \]
\end{definition}

The TRP quantifies the urgency of a request $r$ to be batched compared to admitted requests with the strictest TPOT SLOs. Note that the TRP of a request adaptively responds to changing workloads. Therefore, each request earns credits at its TRP. Accumulating sufficient credit ($\ge 1.0$) grants a request to be batched in the next processing batch. Its credit is then decremented by 1.0, representing the consumption of one processing opportunity. Formally, in each batching step $t$, the following actions are taken:
    \begin{itemize}[leftmargin=*]
        \item \textbf{Credit Earning:} For every request $r \in R(t)$, its credit is updated based on its TRP rate:
        $$ C_r(t) \leftarrow C_r(t) + TRP(r) $$
        where $C_r(t)$ is the credit of request $r$ at step $t$, which is initialized as 0.
        \item \textbf{Batch Selection:} The batch $\mathcal{B}(t)$ is formed by including all requests whose credit, after accumulation, is greater than or equal to the threshold:
        $$ \mathcal{B}(t) = \{r \in R(t) \mid C_r(t) \ge 1.0\} $$
        \item \textbf{Credit Debit:} For every request $r$ included in $\mathcal{B}(t)$, its credit is decremented by 1.0:
        $$ \text{If } r \in \mathcal{B}(t), \text{ then } C_r(t) \leftarrow C_r(t) - 1.0 $$
    \end{itemize}

This mechanism ensures that over many steps, the frequency of a request $r$ being batched will converge towards its TRP rate.

\myline{VBS-based Admission Control.}
When a new request $r$ arrives, to guarantee TPOT SLOs, an intuitive approach is to admit requests into the running batch if admitting it will neither violate its own TPOT SLO nor cause other running requests to violate their TPOT SLOs. However, since credit-based batching causes some requests to skip some execution iterations as mentioned above, directly using the number of running requests as batch size overestimates the actual system load. Since the request $r$ can earn a  $TRP(r)$ opportunity to be batched in each iteration if admitted, it can be regarded as a virtual $TRP(r)$ request. Let $R(t)' = R(t) \cup \{r\}$, we can project the actual load of the system as the sum of the TRP of all requests, which we denote as virtual batch size (VBS):

\begin{equation}
    \text{VBS}(R') = \sum_{r \in R'} TRP(r)
\end{equation}

The request $r$ is admitted to the running queue at step $t$ if adding it would not cause the estimated TPOT to exceed the minimum TPOT SLO of the running requests set $R'$:
$$ 
\quad \text{EstimatedTPOT}(VBS(R'), L_{avg}(R')) \leq {\min_{r' \in R'} S_{TP}(r')} $$
The mechanism drives the admitted requests toward their TPOT SLOs. Because the predictor and analytic models are imperfect, \system{} targets high statistical attainment rather than hard guarantees, and rejecting requests estimated to violate their TPOT SLOs bounds the impact of mispredictions.

\definecolor{myPurple}{rgb}{0.5, 0.0, 0.8}
\SetCommentSty{mycommentstyle}
\newcommand{\mycommentstyle}[1]{\textcolor{myPurple}{\textit{#1}}}

\begin{algorithm}[t]
    \caption{TPOT Guarantee Mechanism}
    \label{alg:tpot_guarantee_mechanism}
    \DontPrintSemicolon
    \SetAlgoLined
    \KwIn{LLM model $M$, Sequence Length Predictor $P$, LDF Sorted Waiting Queue $W$, Running Queue $R$}
    \KwOut{Batched requests set $B$}
    
    \While{True}{
        $B \gets \emptyset$\;
        \ForEach{$w \in W$}{ 
            $R' \gets R \cup \{w\}$\;
            \If{$\text{EstimatedTPOT}(VBS(R'), L_{\text{avg}}(R')) \leq \min_{r \in R'} S_{TP}(r)$}{ 
                $B \gets B \cup \{w\}$, $W \gets W \setminus \{w\}$\,
                , $R \gets R'$ \tcp*{Admission Control}
            }
        }

        \ForEach{$r \in R$}{
            $C_r(t) \gets C_r(t) + TRP(r)$ \tcp*{Credit Earning}
            \If{$C_r(t) \ge 1.0$}{
                $B \gets B \cup \{r\}$, $C_r(t) \gets C_r(t) - 1.0$ \tcp*{Batch Selection and Credit Debit}
            }
        }
    }
\end{algorithm}

\section{Experiments} \label{sec:exp}
We evaluate \system{} against state-of-the-art baselines and analyze the contribution of each component.

\subsection{Experimental Setup}
\label{subsec:exp_setup}
\myline{Testbed.} We use a server with 4 NVIDIA A100 GPUs (80GB each), connected by NVLink within pairs (GPU0-GPU1, GPU2-GPU3) and PCIe/NUMA across pairs.

\myline{Serving Models.} We use Meta Llama-3.1 8B \citep{llama3} and Google Gemma-2 27b \citep{gemma2} as serving models, in FP16/BF16 precision. The 8B model runs on a single GPU, and the 27B model runs on 4 GPUs with tensor parallelism \citep{megatron-lm}.

\myline{Workloads.} We evaluate on the ShareGPT \citep{ShareGPT} and LMSYS-Chat-1M \citep{lmsyschat1m} datasets, widely used collections of real-world conversations. To stress heterogeneous serving, following prior works \citep{Adaserve,pddisaggreate} we consider 6 SLO categories, summarized in Table \ref{tab:dataset_summary}: Category 1 has tight TTFT and TPOT (e.g., code generation); Categories 2-3 have tight TPOT and looser TTFT (e.g., tool calls); Categories 4-5 have loose TPOT and tight TTFT (e.g., reading-speed chatbots); Category 6 is loose on both (e.g., summarization). For the 27B model we loosen the constraints to account for the larger model size.

\begin{table}[h]
  \centering
  \small
  \caption{SLO categories for different model sizes.}
  \label{tab:dataset_summary}
\begin{tabular}{llcccccc}
\toprule
\multirow{2}{*}{\textbf{Model}} & \multirow{2}{*}{\textbf{Metric}} & \multicolumn{6}{c}{\textbf{Category}} \\
 &  & \textbf{1} & \textbf{2} & \textbf{3} & \textbf{4} & \textbf{5} & \textbf{6} \\ \midrule
\multirow{2}{*}{Llama-3.1 8B} & TTFT (s) & 0.5 & 2 & 3 & 0.5 & 1 & 7.5 \\
 & TPOT (ms) & 30 & 30 & 30 & 50 & 50 & 50 \\ \hline
\multirow{2}{*}{Gemma-2 27B} & TTFT (s) & 1.0 & 4 & 6 & 1.0 & 2 & 15 \\
 & TPOT (ms) & 60 & 60 & 60 & 100 & 100 & 100 \\
\bottomrule
\end{tabular}
\end{table}

\myline{Baselines.} We compare our method with the following baselines:

\begin{itemize}[leftmargin=*]
\item \textbf{vLLM} \citep{vllm}: A state-of-the-art LLM serving system that uses a throughput-oriented scheduling strategy. The vLLM scheduler prioritizes prefills. 

\item \textbf{S3} \citep{s3}: An LLM inference system that predicts the output sequence length and employs the shortest job first scheduling. Since S3 is implemented on basic LLM inference without techniques like PagedAttention, we re-implement its core scheduling strategy within vLLM for a fair comparison.
\item \textbf{Mooncake} \citep{MoonCake}: A production-grade LLM serving platform that employs an early-rejection-based admission control for SLO guarantees. We integrate this mechanism into vLLM.
\end{itemize}

\subsection{QPS-Scaling SLO Attainment} \label{subsec:qps_scaling_slo_attainment}
\begin{figure*}[t]
  \centering
  \includegraphics[scale=0.098]{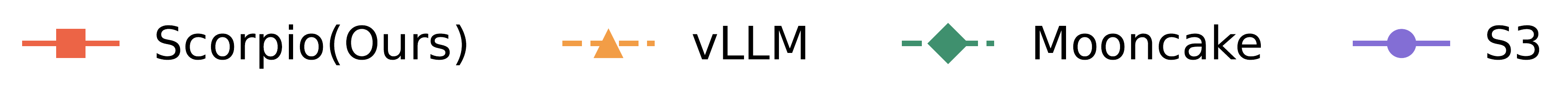} \\
  \begin{subfigure}[b]{0.24\textwidth}
    \centering
    \includegraphics[scale=0.098]{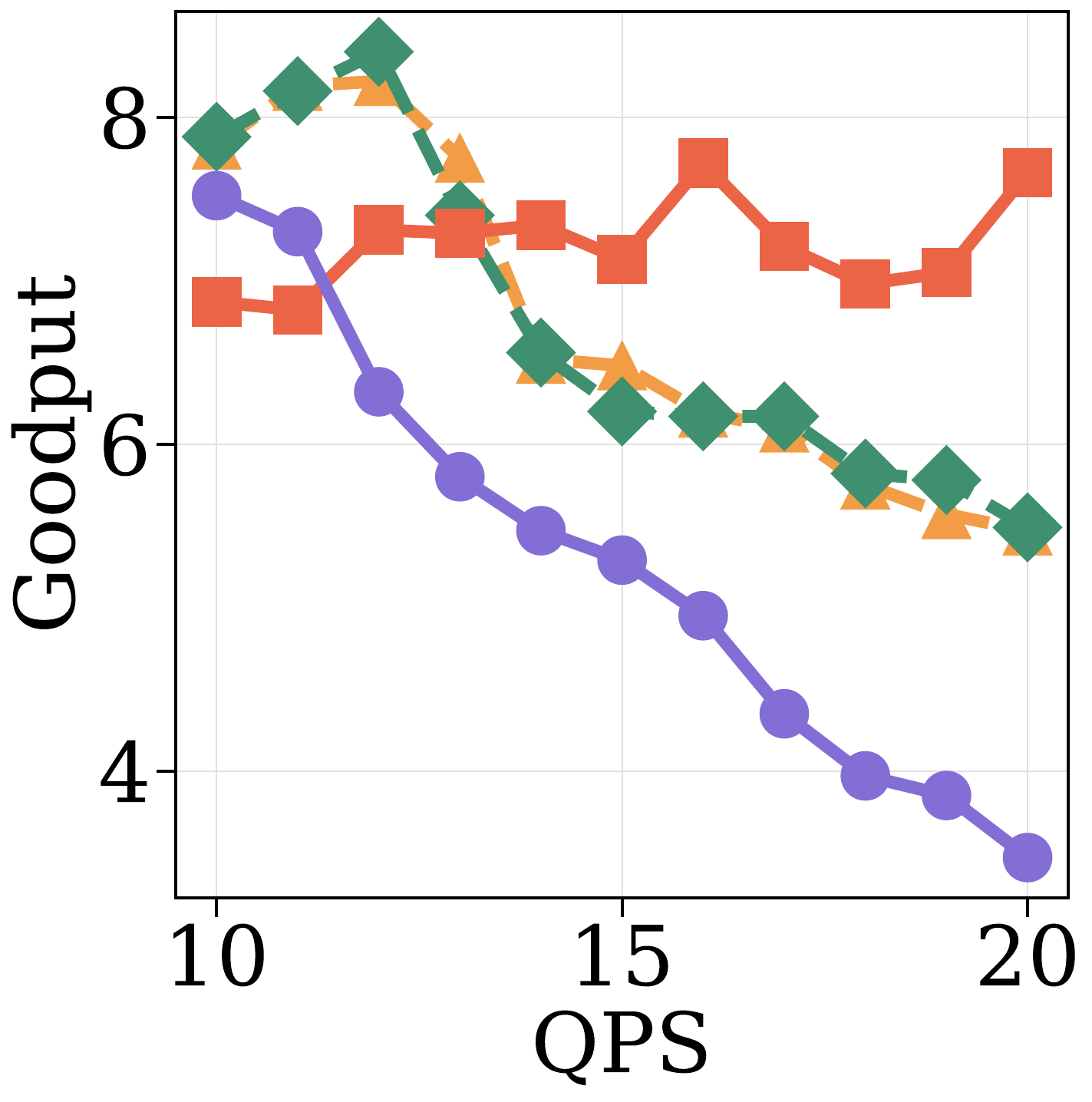}
  \end{subfigure}
  \hfill
  \begin{subfigure}[b]{0.24\textwidth}
    \centering
    \includegraphics[scale=0.098]{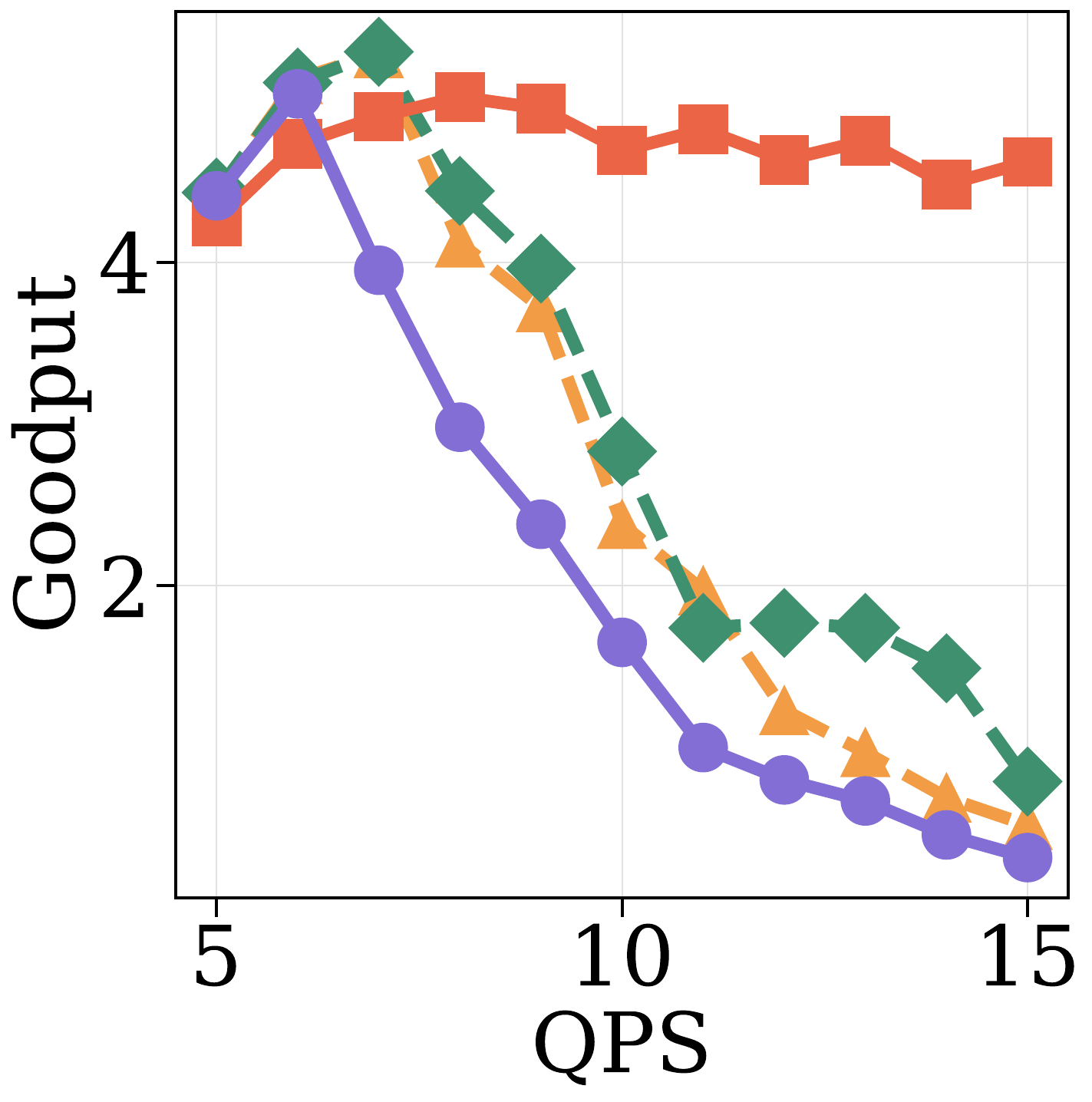}
  \end{subfigure}
  \hfill
  \begin{subfigure}[b]{0.24\textwidth}
    \centering
    \includegraphics[scale=0.098]{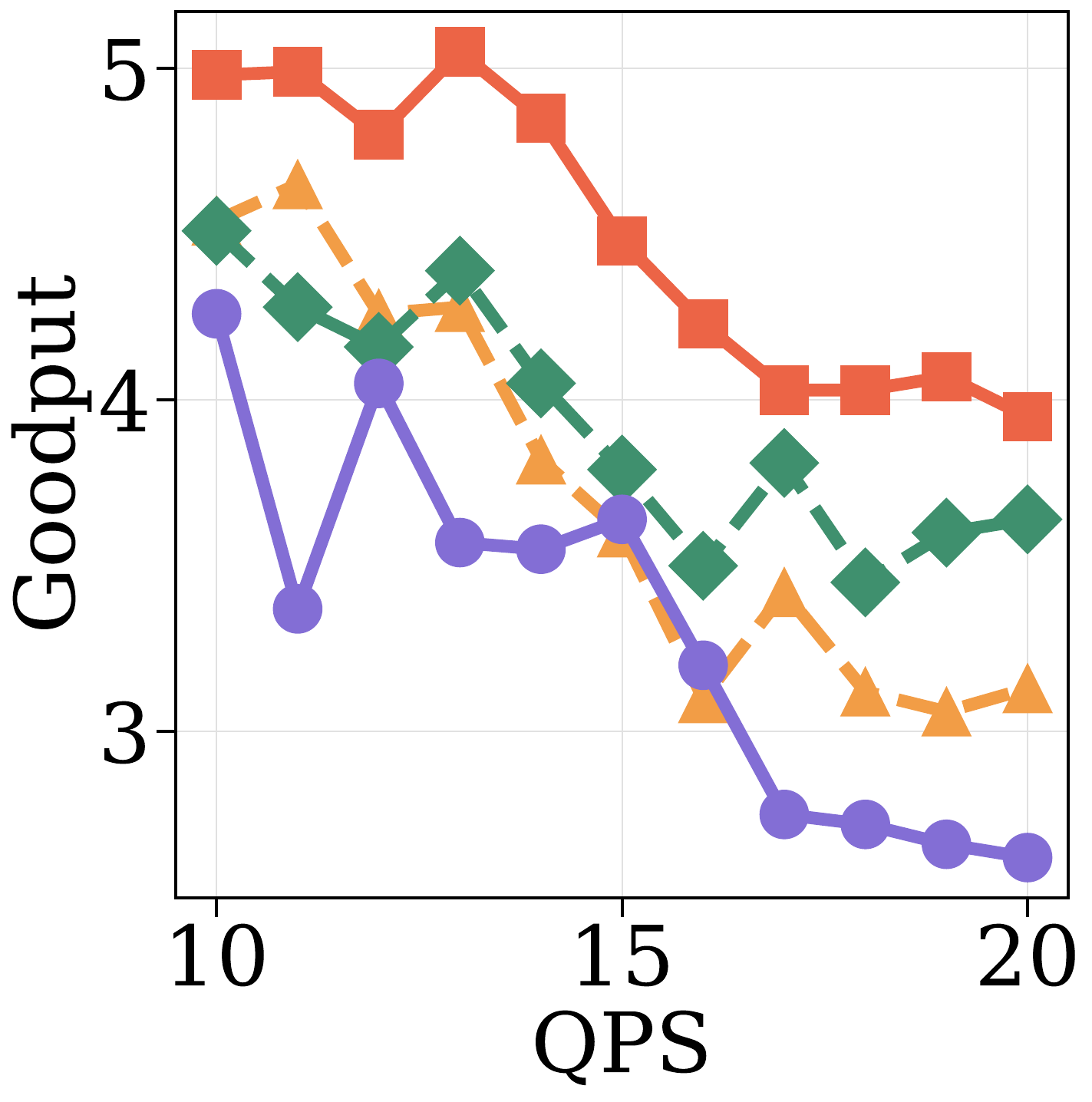}
  \end{subfigure}
  \hfill
  \begin{subfigure}[b]{0.24\textwidth}
    \centering
    \includegraphics[scale=0.098]{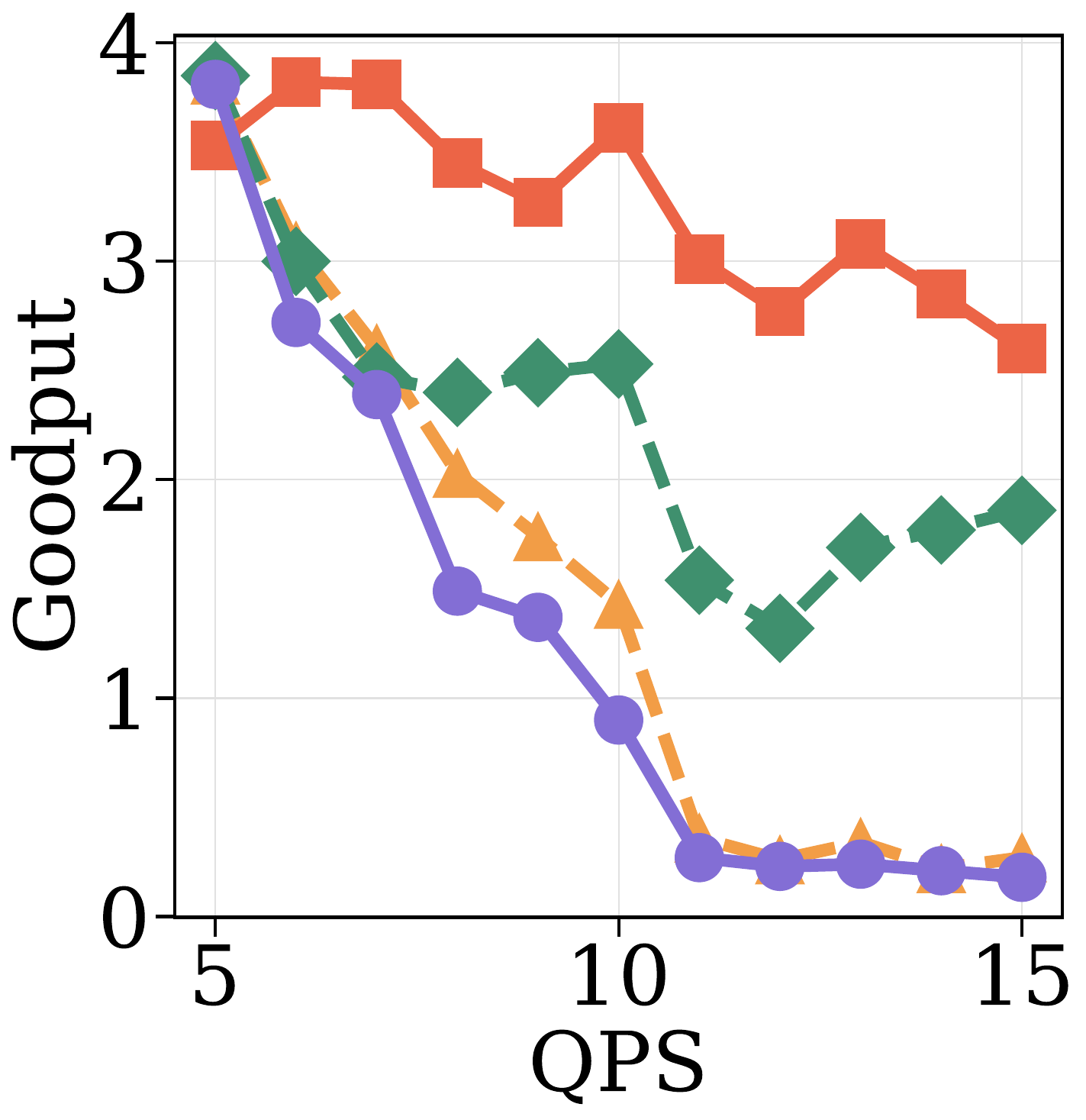}
  \end{subfigure}
  \hfill
  \begin{subfigure}[b]{0.24\textwidth}
    \centering
    \includegraphics[scale=0.098]{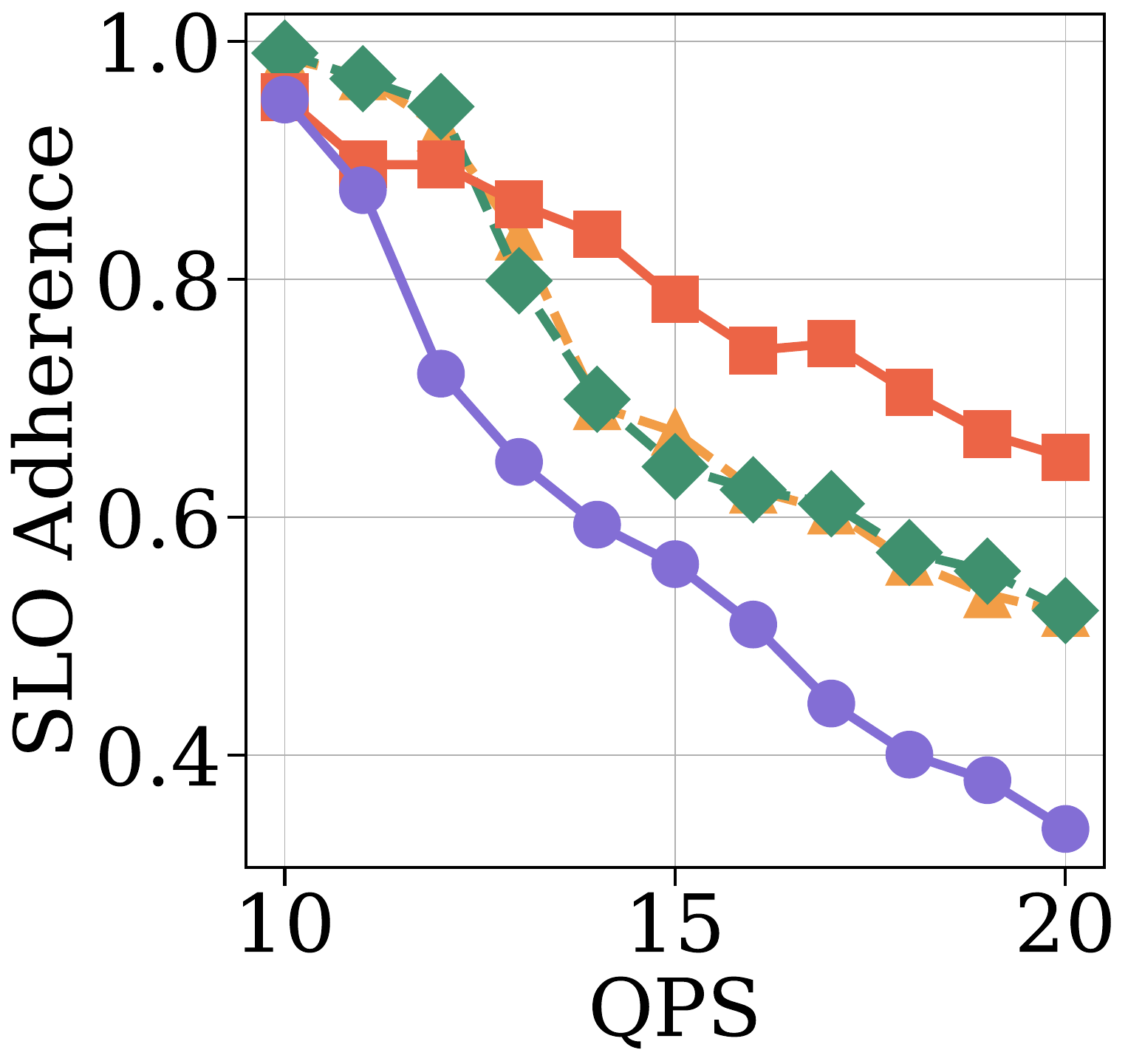}
    \caption{Llama3-8B\\LMSYS}
  \end{subfigure}
  \hfill
  \begin{subfigure}[b]{0.24\textwidth}
    \centering
    \includegraphics[scale=0.098]{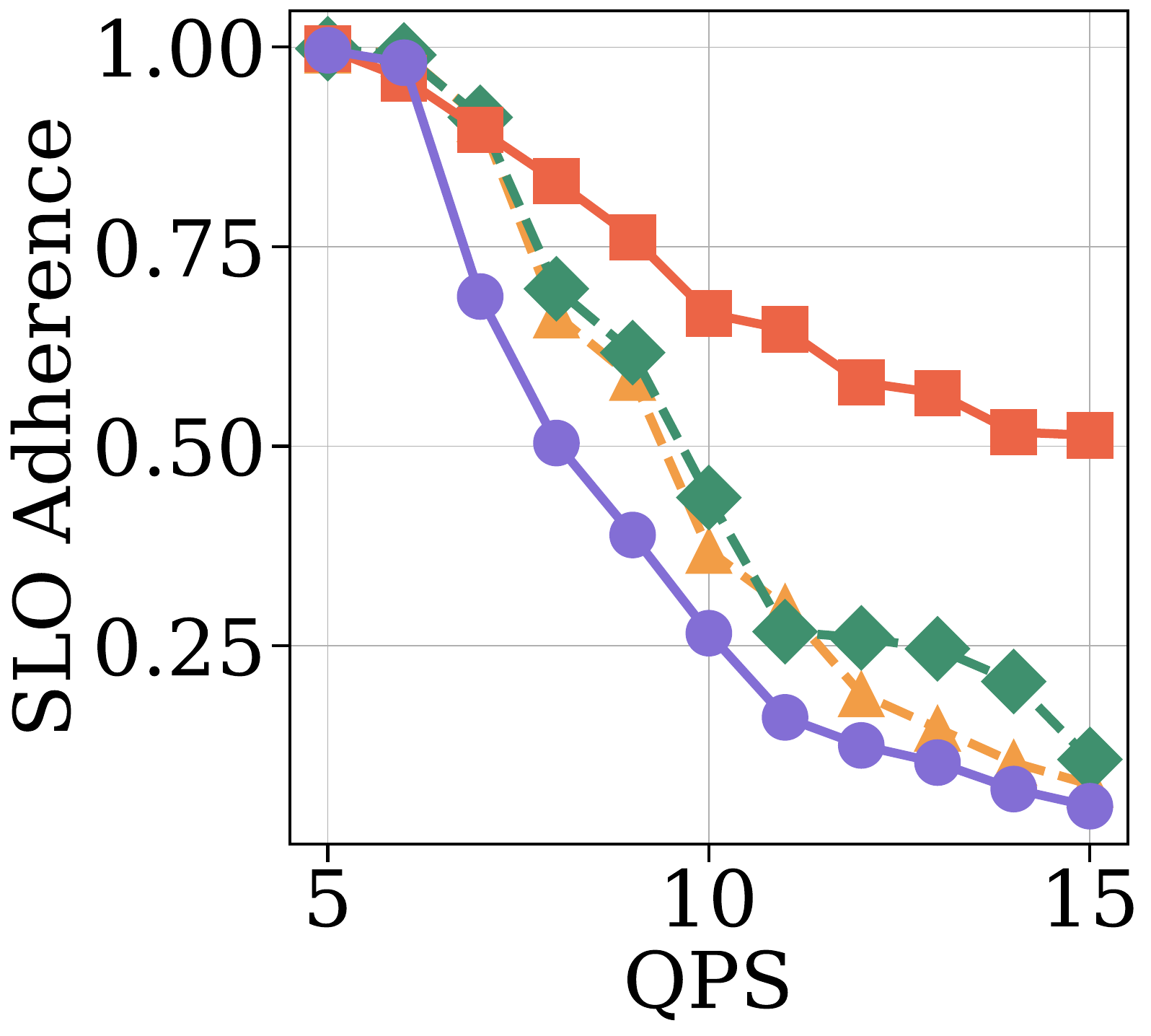}
    \caption{Llama3-8B\\ShareGPT}
  \end{subfigure}
  \hfill
  \begin{subfigure}[b]{0.24\textwidth}
    \centering
    \includegraphics[scale=0.098]{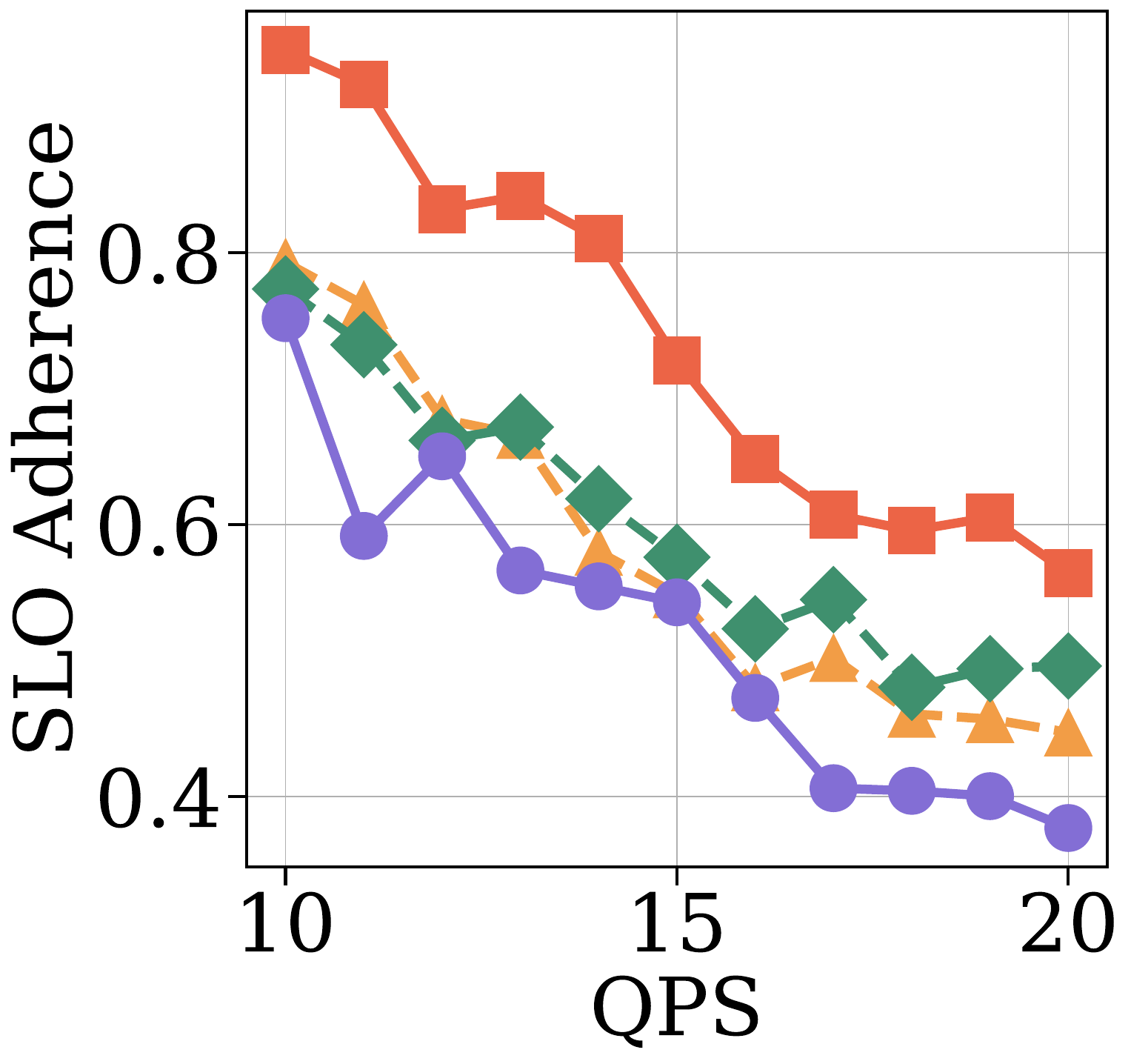}
    \caption{Gemma-27B\\LMSYS}
  \end{subfigure}
  \hfill
  \begin{subfigure}[b]{0.24\textwidth}
    \centering
    \includegraphics[scale=0.098]{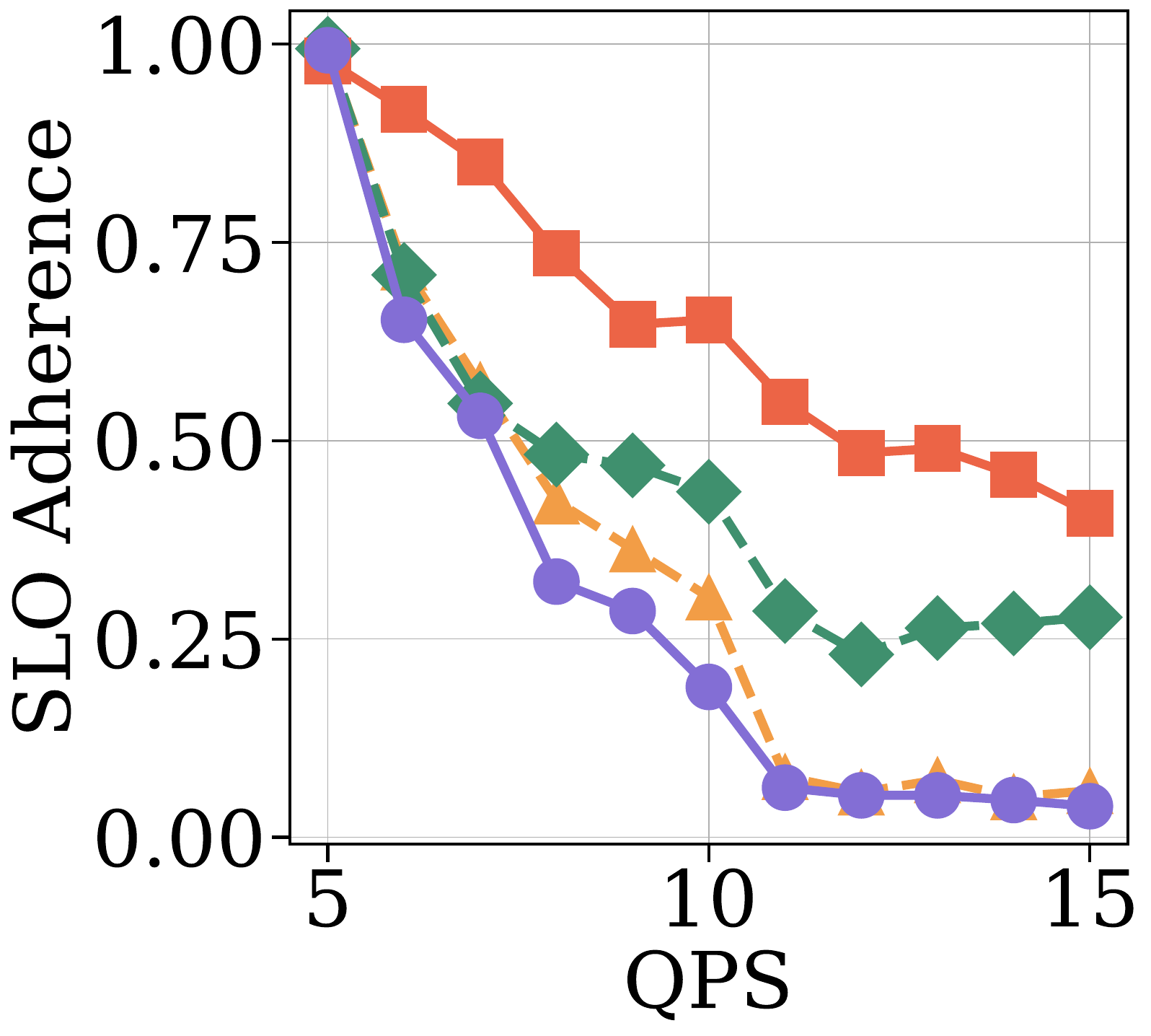}
    \caption{Gemma-27B\\ShareGPT}
  \end{subfigure}
  \caption{Impact of different scheduling strategies on goodput and SLO adherence vs QPS.}
  \label{fig:exp1_qps}
\end{figure*}

We compare \system's goodput and SLO attainment rate against baselines on ShareGPT and LMSYS-Chat-1M with heterogeneous SLOs, under varying request arrival rates \citep{Adaserve,pddisaggreate}. As shown in Figure \ref{fig:exp1_qps}, \system{}  achieves higher goodput and SLO adherence than the baselines, especially at high QPS. For example, at a QPS of 15, it yields up to 8.8-14.4$\times$ higher goodput and 40.7-46.5\% higher SLO adherence than baselines, demonstrating robust burst handling through its SLO-oriented scheduling. In contrast, vLLM's greedy admission leads to severe TPOT violations. S3's output-length ranking and Mooncake's strict rejection both exhibit suboptimal SLO adherence. At low load the system is not the bottleneck and all methods meet most SLOs, so the small overhead of co-locating the predictor (quantified in \S\ref{subsec:effectiveness_analysis}) slightly reduces \system's goodput in the regime where scheduling matters least. On long-input and long-output workloads (1K-2K tokens), \system{} sustains consistent gains over baselines, in line with the main results.

  \subsection{Real-World Trace Serving}
  \label{subsec:real_world_trace_serving}
\begin{figure*}[t]
  \centering
  \includegraphics[scale=0.098]{misc/legend/slo_goodput_legend.pdf} \\
  \begin{subfigure}[b]{0.24\textwidth}
    \centering
    \includegraphics[scale=0.098]{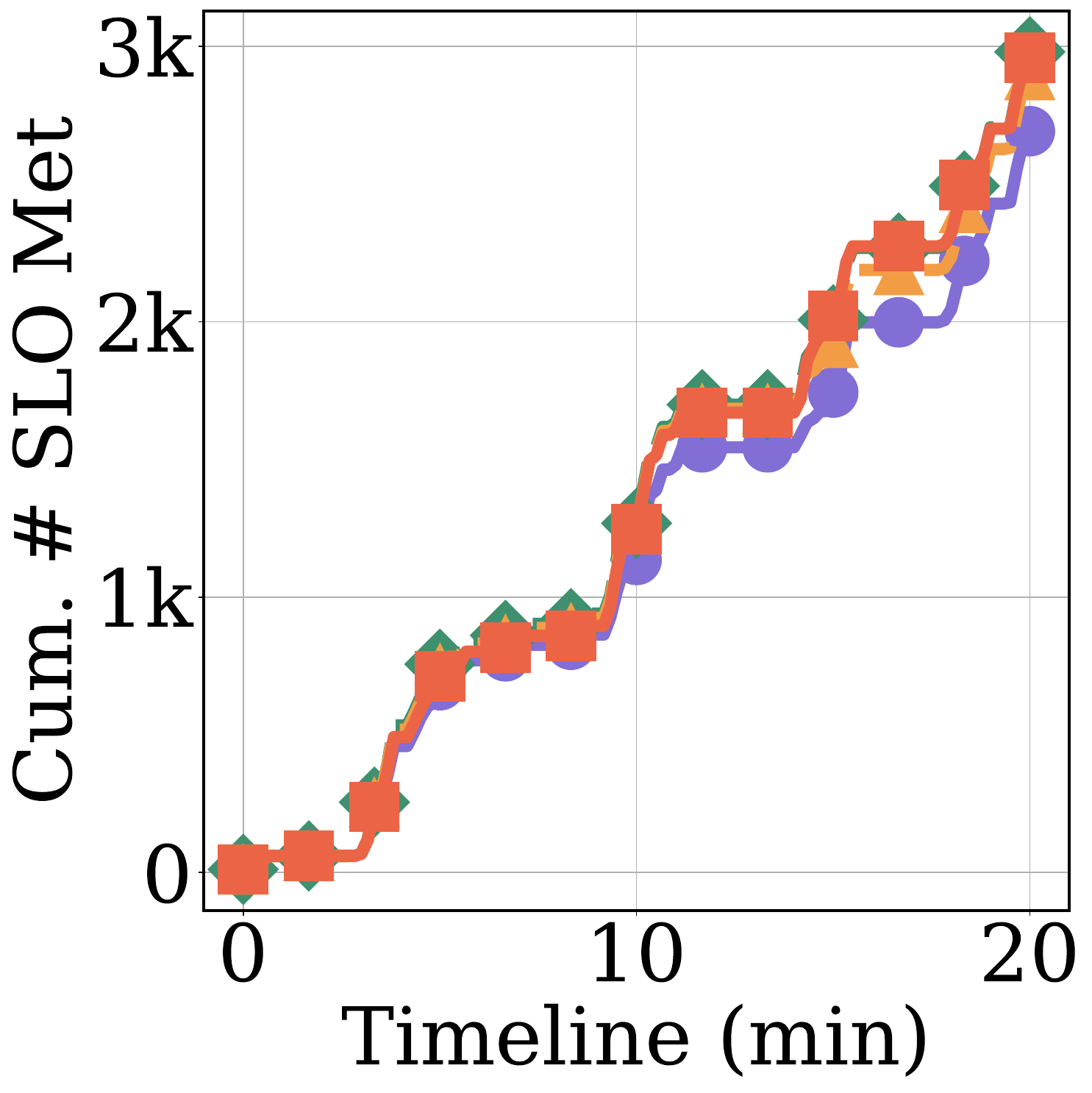}
    \caption{Llama3-8B\\LMSYS}
    \label{subfig:trace_a}
  \end{subfigure}
  \hfill
  \begin{subfigure}[b]{0.24\textwidth}
    \centering
    \includegraphics[scale=0.098]{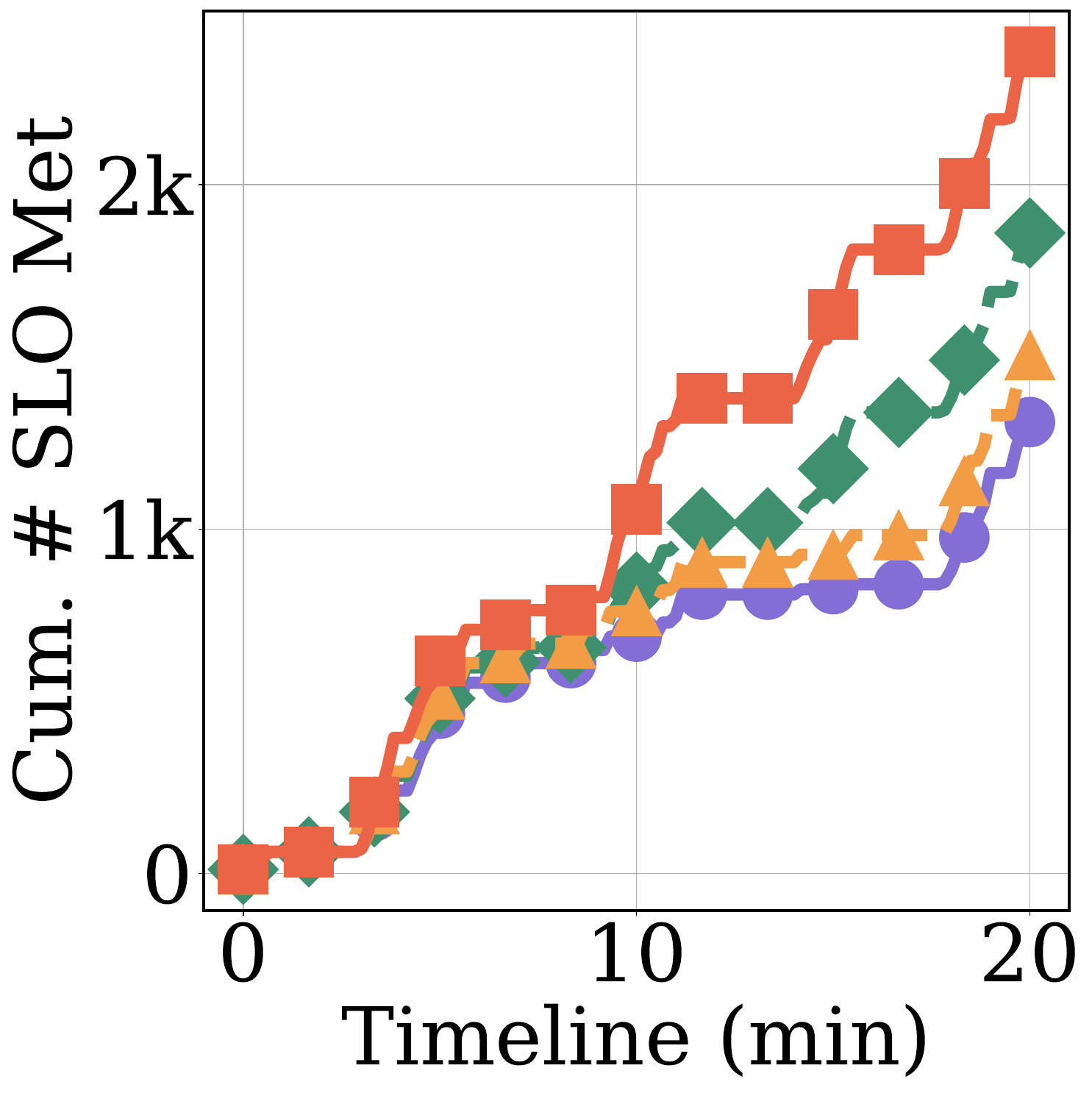}
    \caption{Llama3-8B\\ShareGPT}
    \label{subfig:trace_b}
  \end{subfigure}
  \hfill
  \begin{subfigure}[b]{0.24\textwidth}
    \centering
    \includegraphics[scale=0.098]{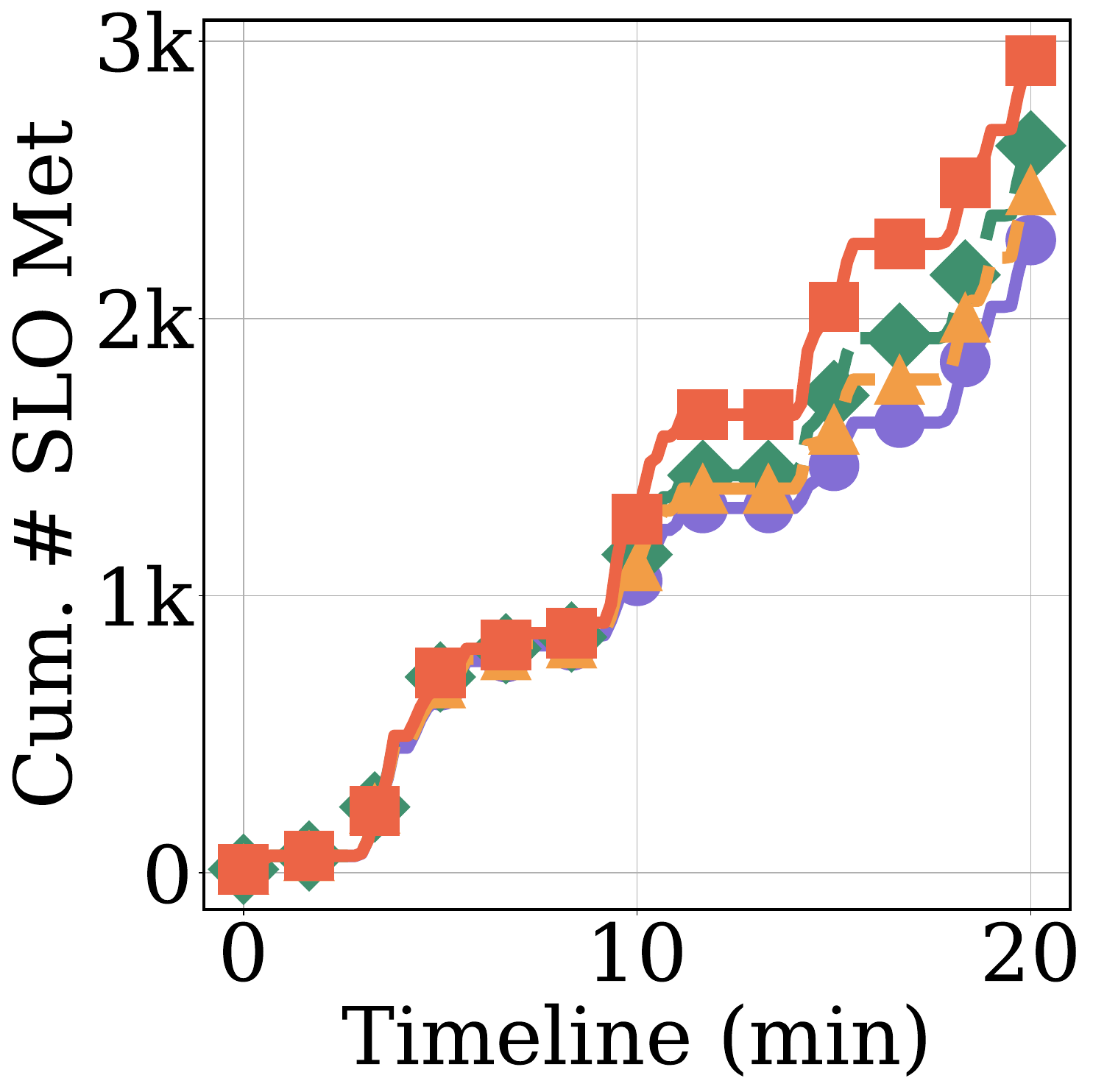}
    \caption{Gemma-27B\\LMSYS}
    \label{subfig:trace_c}
  \end{subfigure}
  \hfill
  \begin{subfigure}[b]{0.24\textwidth}
    \centering
    \includegraphics[scale=0.098]{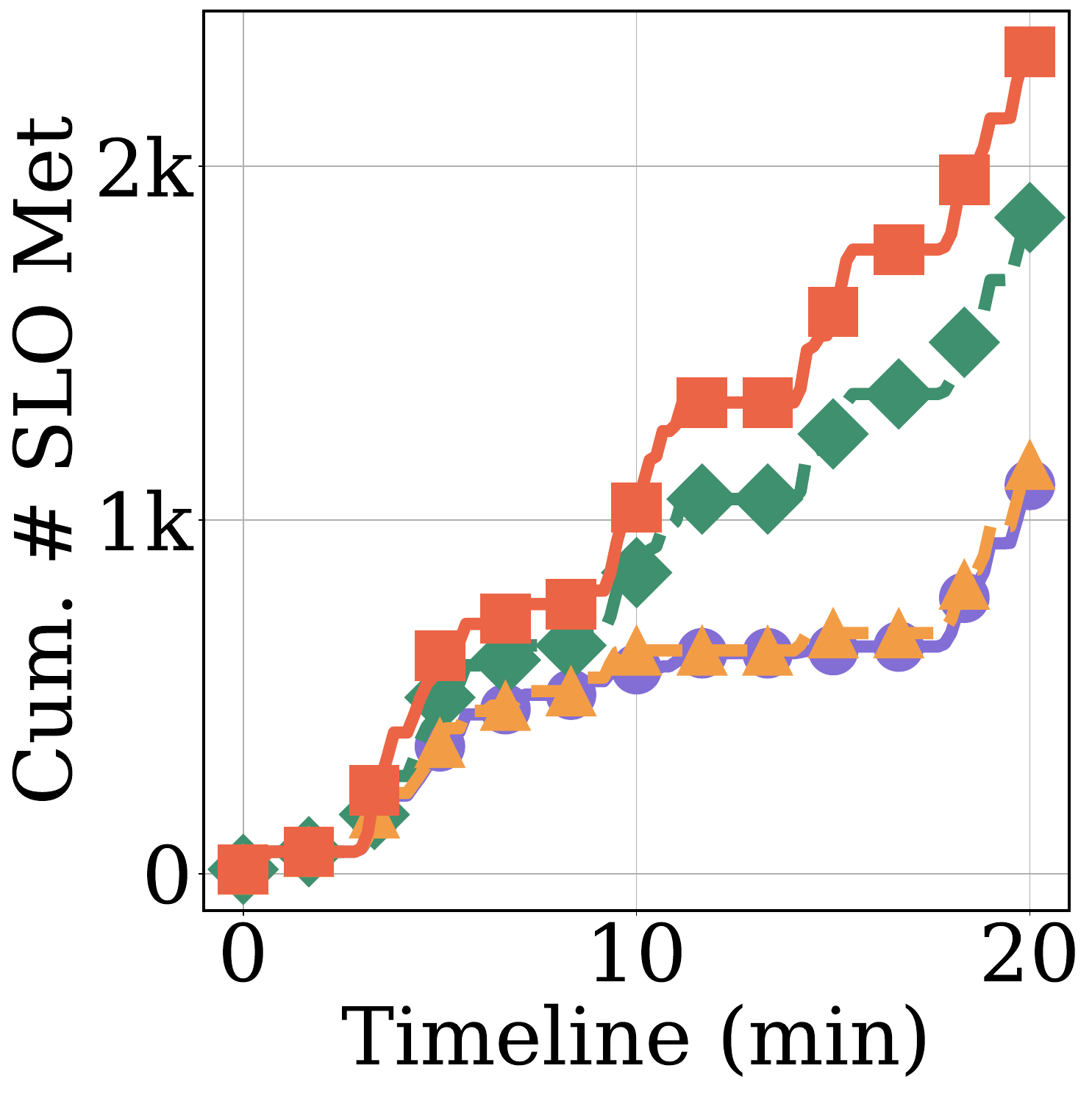}
    \caption{Gemma-27B\\ShareGPT}
    \label{subfig:trace_d}
  \end{subfigure}
  \caption{Cumulative number of SLO-met requests over time on the LMSYS and ShareGPT dataset.}
  \label{fig:trace}
\end{figure*}

  We further evaluate \system{} using 20-minute real-world traces \citep{AzureTrace}; the 20-min Azure trace mixes bursty and light load, peaking above 60 req/s. Figure~\ref{fig:trace} reports the cumulative number of SLO-met requests for Llama3-8B and Gemma2-27B models across both datasets. Across all settings, \system{} achieves the highest cumulative SLO-met counts. For example, \system{} achieves a 1.25, 2.01, and 2.11$\times$ higher SLO-met requests compared with Mooncake, vLLM, and S3, respectively. Throughput-oriented baselines like vLLM exhibit noticeable slowdowns during traffic spikes (e.g., minutes 3-5 and 9-11), caused by uniform batching and greedy admission, leading to widespread TPOT violations. Mooncake (Figure~\ref{fig:trace}b,d) alleviates some pressure via early rejections, but its SLO gains taper off with a too strict admission control mechanism and lack of consideration for the heterogeneous SLOs. S3's length ranking also exhibits suboptimal SLO attainment. In contrast, \system{} steadily maintains a lead in cumulative SLO-met  requests. By combining TTFT Guard and TPOT Guard, \system{} effectively handles the heterogeneous SLOs.

\begin{figure*}[t]
  \centering
  \includegraphics[scale=0.098]{misc/legend/slo_goodput_legend.pdf} \\
  \begin{subfigure}[b]{0.24\textwidth}
    \centering
    \includegraphics[scale=0.098]{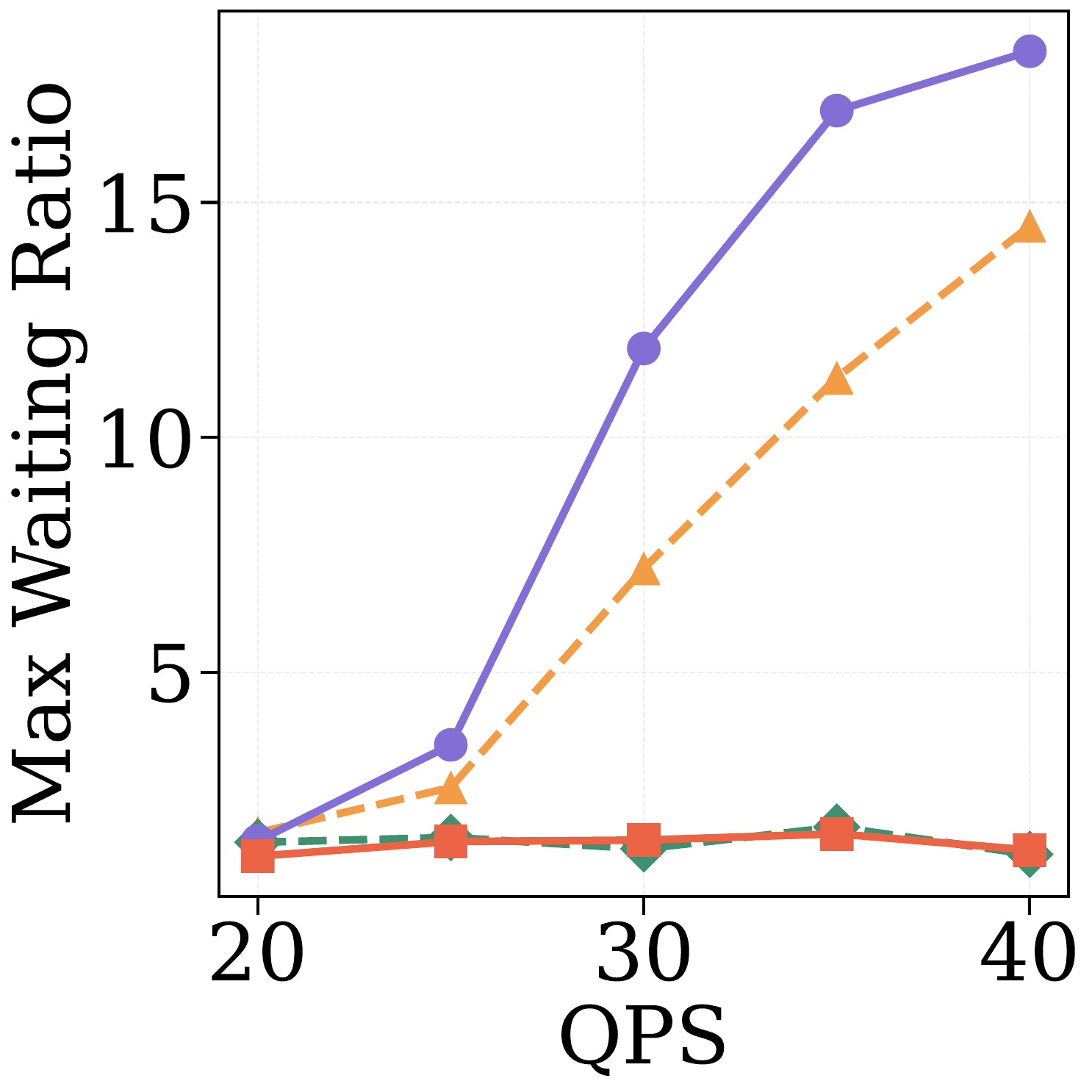}
    \caption{Llama3-8B\\LMSYS}
    \label{subfig:starv_a}
  \end{subfigure}
  \hfill
  \begin{subfigure}[b]{0.24\textwidth}
    \centering
    \includegraphics[scale=0.098]{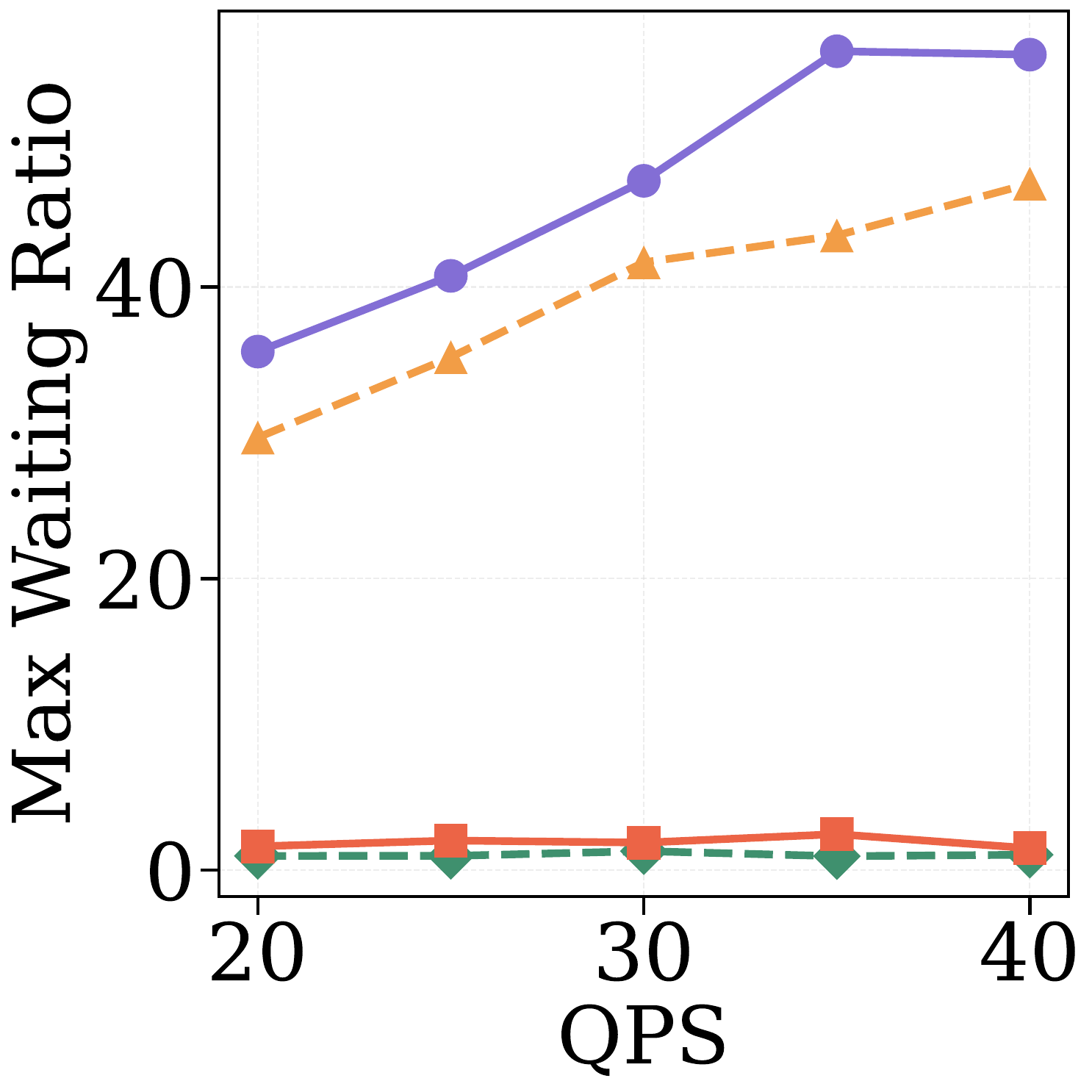}
    \caption{Llama3-8B\\ShareGPT}
    \label{subfig:starv_b}
  \end{subfigure}
  \hfill
  \begin{subfigure}[b]{0.24\textwidth}
    \centering
    \includegraphics[scale=0.098]{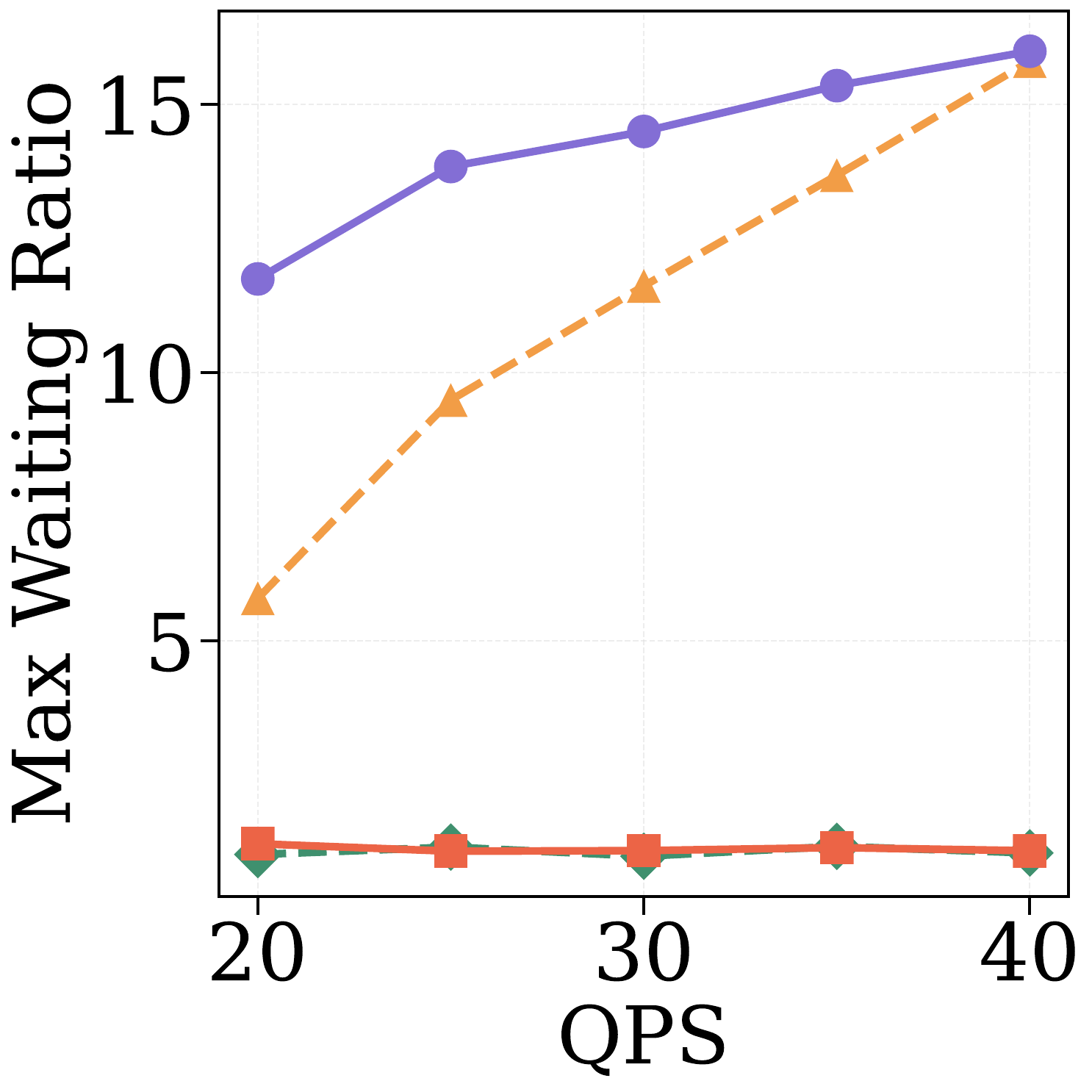}
    \caption{Gemma-27B\\LMSYS}
    \label{subfig:starv_c}
  \end{subfigure}
  \hfill
  \begin{subfigure}[b]{0.24\textwidth}
    \centering
    \includegraphics[scale=0.098]{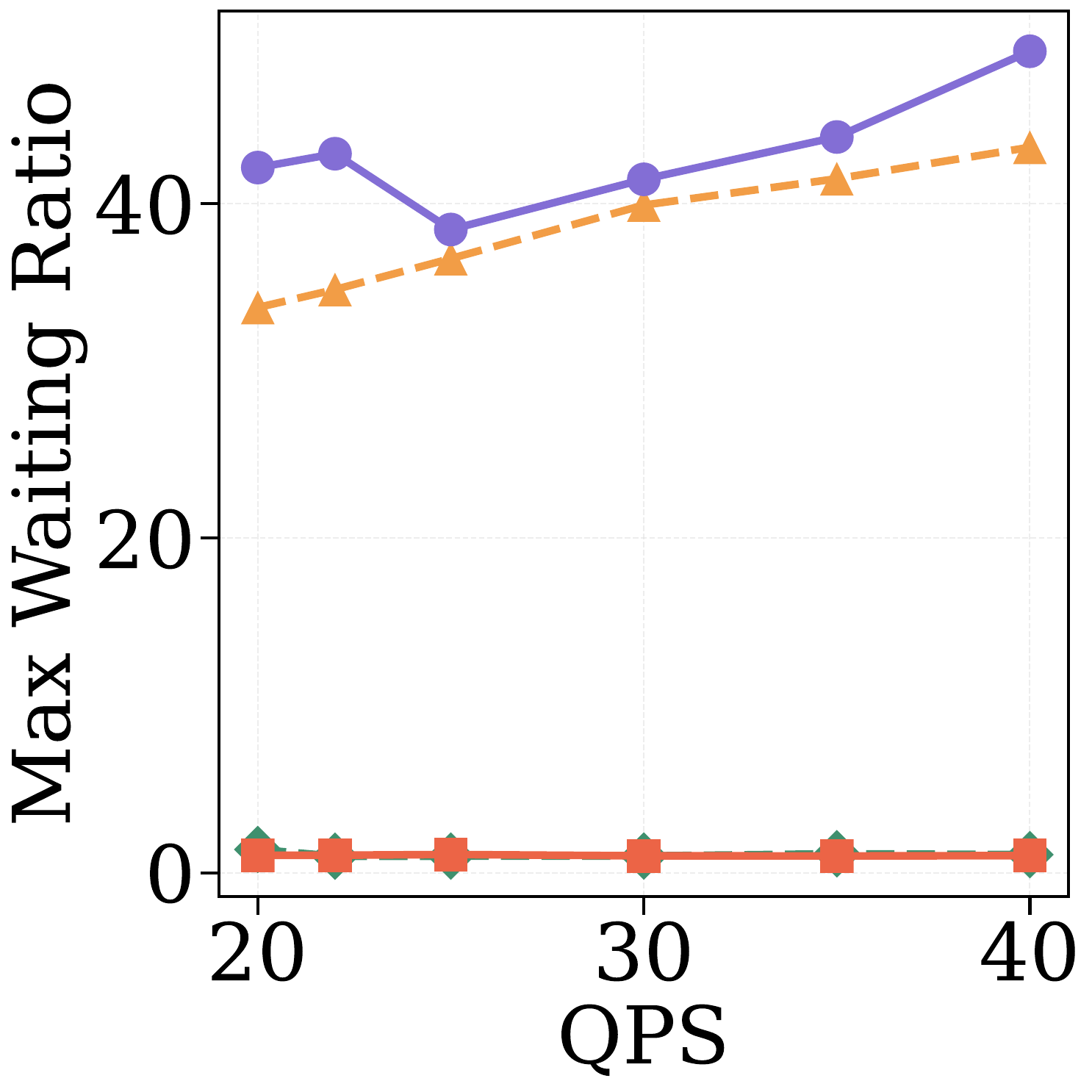}
    \caption{Gemma-27B\\ShareGPT}
    \label{subfig:starv_d}
  \end{subfigure}
  \caption{Starvation prevention on the impact of different scheduling strategies.}
  \label{fig:starvation}
\end{figure*}

\subsection{Effectiveness Analysis}
\label{subsec:effectiveness_analysis}

\begin{figure}[t]
  \centering
  \includegraphics[scale=0.098]{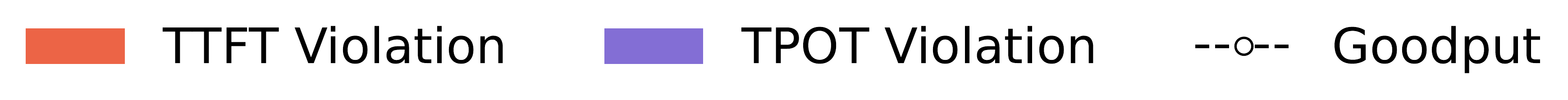} \\
  \begin{subfigure}[b]{0.24\textwidth}
    \centering
    \includegraphics[scale=0.098]{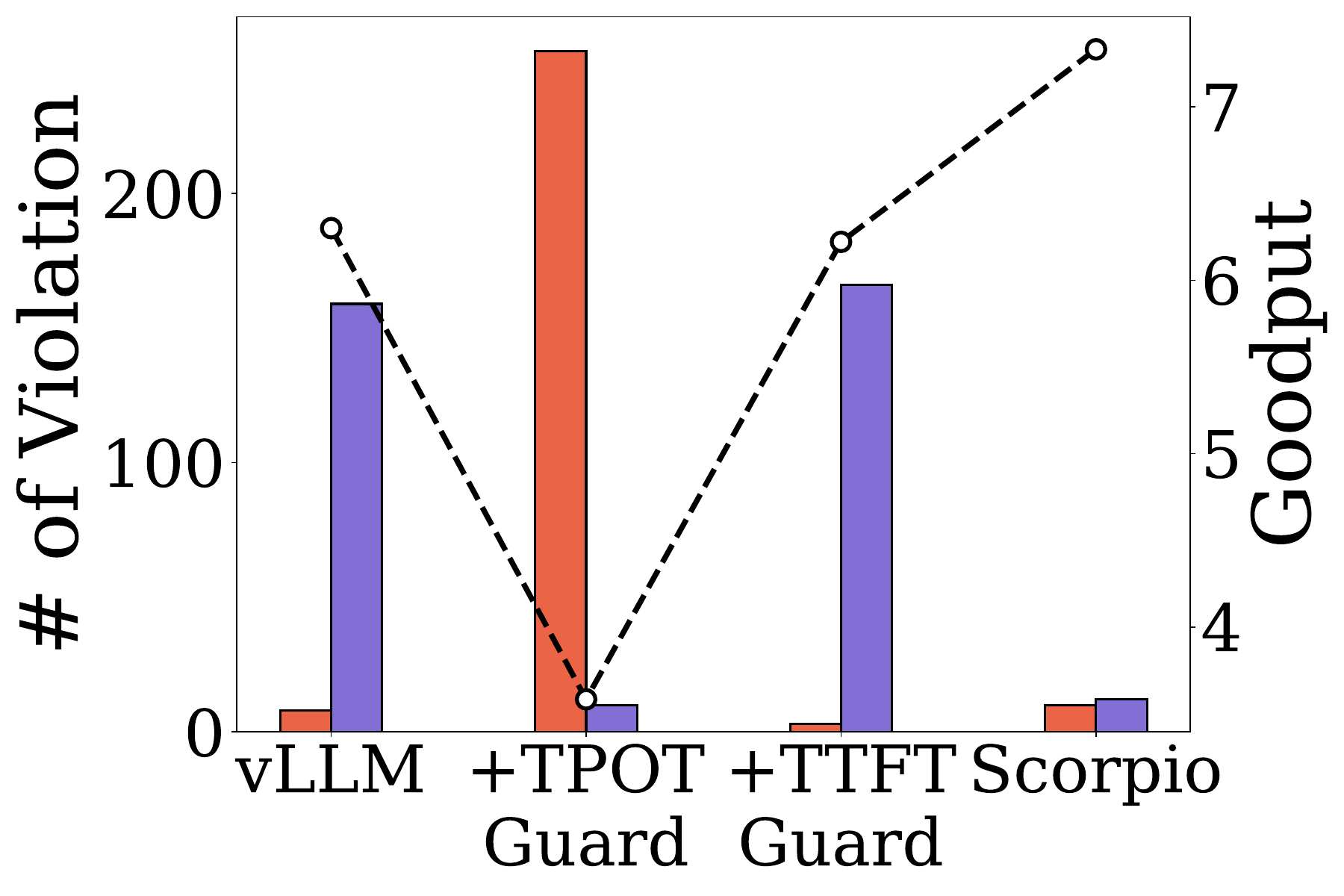}
    \caption{Llama3-8B\\LMSYS}
  \end{subfigure}
  \hfill
  \begin{subfigure}[b]{0.24\textwidth}
    \centering
    \includegraphics[scale=0.098]{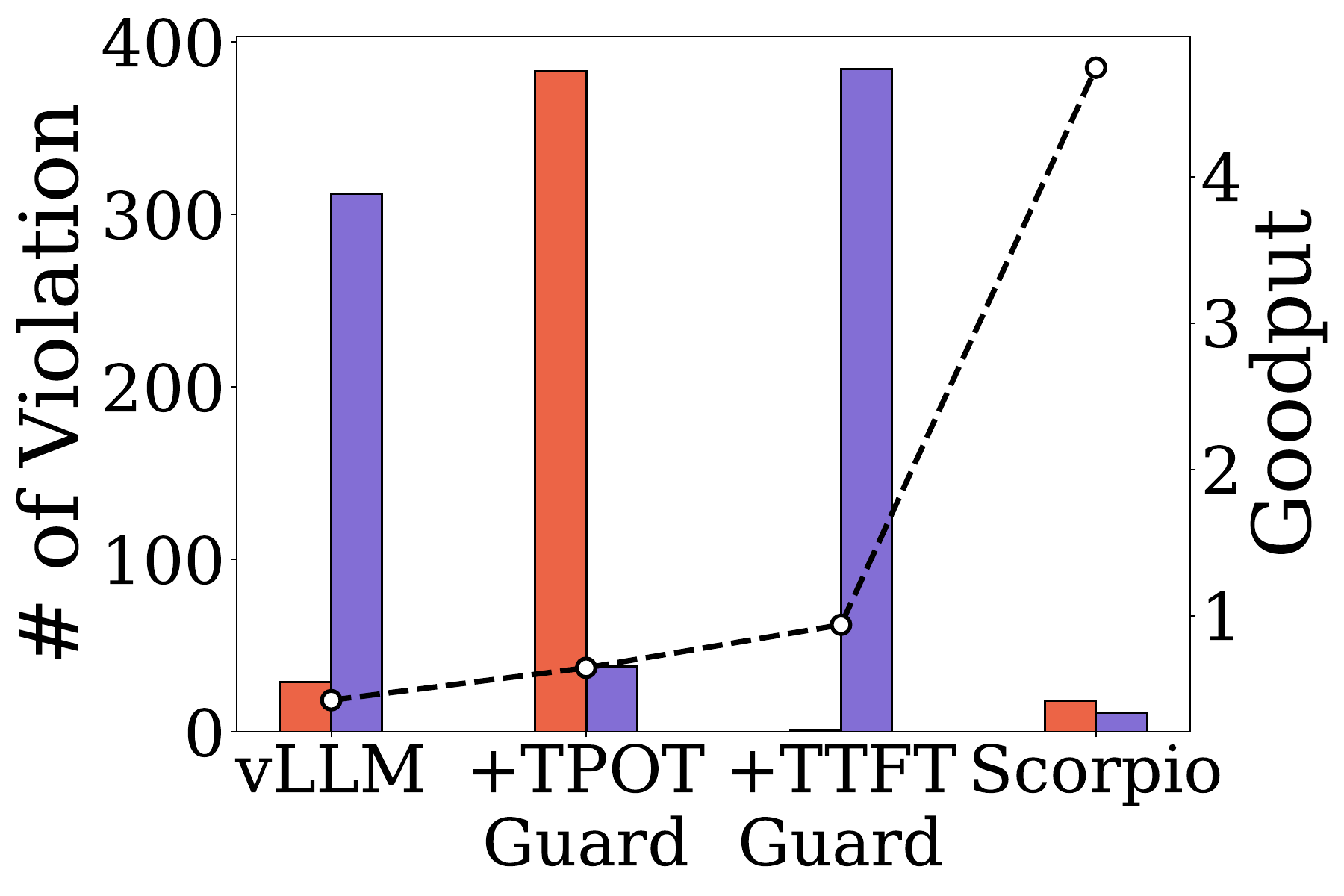}
    \caption{Llama3-8B\\ShareGPT}
  \end{subfigure}
  \hfill
  \begin{subfigure}[b]{0.24\textwidth}
    \centering
    \includegraphics[scale=0.098]{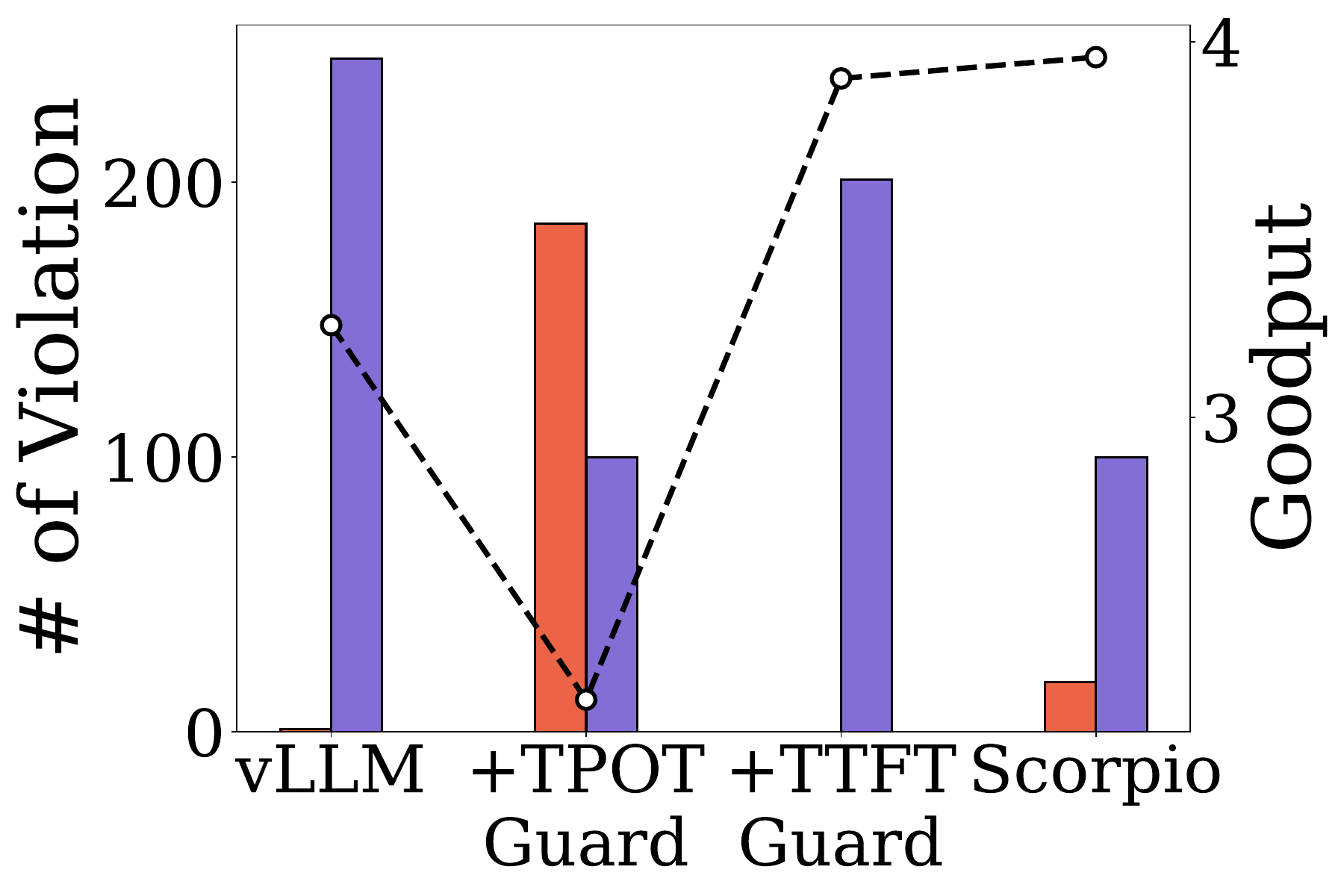}
    \caption{Gemma-27B\\LMSYS}
  \end{subfigure}
  \hfill
  \begin{subfigure}[b]{0.24\textwidth}
    \centering
    \includegraphics[scale=0.098]{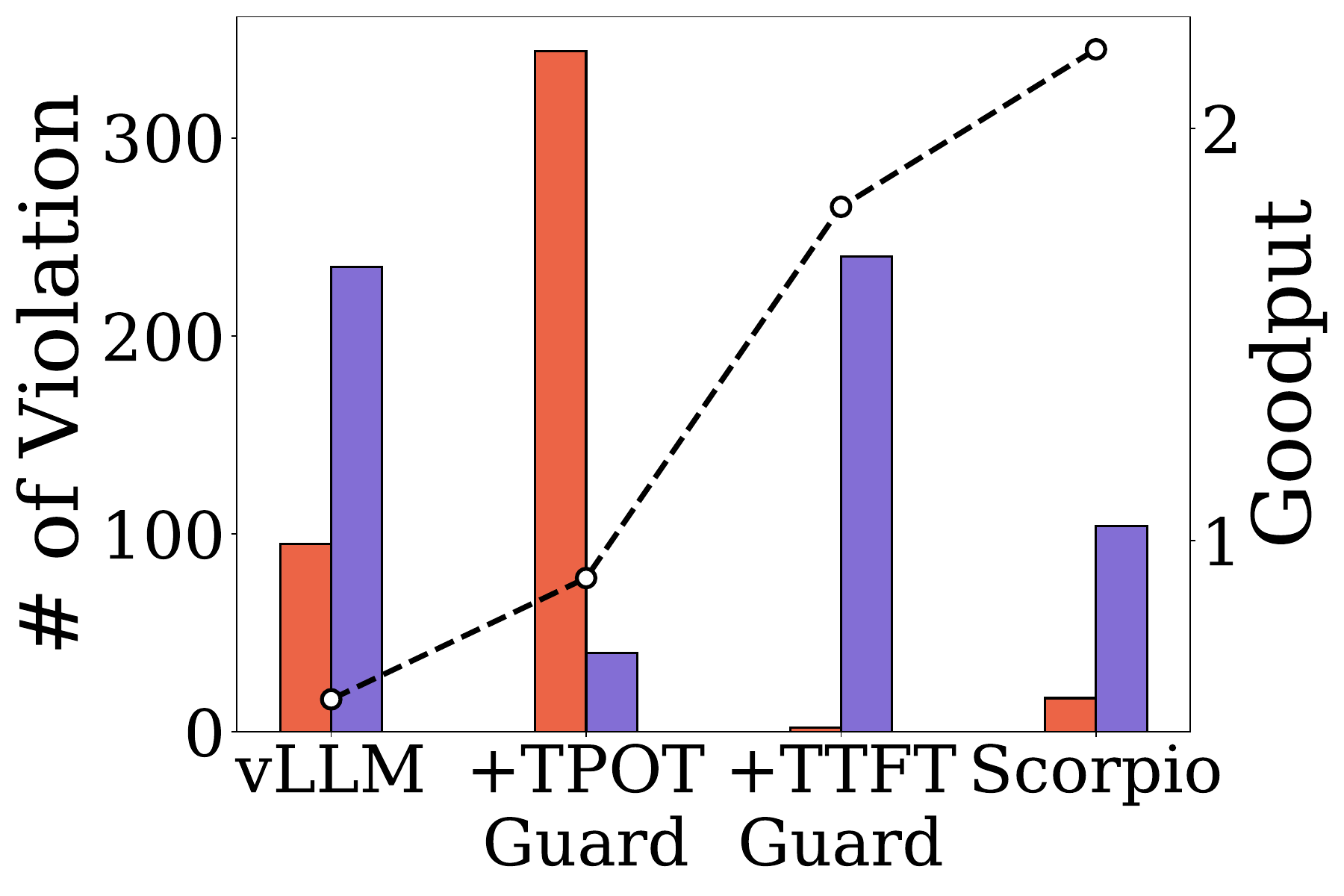}
    \caption{Gemma-27B\\ShareGPT}
  \end{subfigure}
  \caption{Ablation study on the impact of different scheduling strategies.}
  \label{fig:exp3_ablation}
\end{figure}

\myline{Ablation Study.} For this evaluation, we evaluate the core components of \system. As shown in Figure \ref{fig:exp3_ablation}, we add the proposed components incrementally and assess the performance at a representative high-load QPS of 14. We find that 1) without any component, \system's goodput is significantly reduced. 2) Including only the TTFT Guard effectively reduces TTFT violations as intended, but results in severe TPOT violations, and vice versa. These results show the interdependence of the components, highlighting their importance for the overall performance.

\myline{Overhead of \system's Scheduling.} Summing the time of all scheduling sub-components (TTFT Guard and TPOT Guard) when responding to 512 requests, \system{} adds under 1\% scheduling overhead in all four model-dataset settings (0.11\% to 0.17\%). We do not account the time of the predictor since it is shown to be negligible in previous works \citep{s3,learbyrank}.

\myline{Starvation Prevention.} We evaluate the starvation prevention of \system{}, shown in Figure \ref{fig:starvation}. \system{} achieves comparable mean $max\_waiting\_ratio$ with Mooncake, while outperforming other baselines by 1-2 orders of magnitude. These results confirm that \system{} successfully prevents request starvation in heterogeneous SLO environments.

\section{Related Work} \label{sec:related}

\myline{Throughput-oriented serving.} A first line of work pursues high serving throughput by improving GPU utilization. Orca \citep{Orca} and vLLM \citep{vllm} introduce continuous batching and paged attention, which have become de facto building blocks for modern serving stacks \citep{LiInferenceServing} and production engines such as SGLang \citep{sglang} and TensorRT-LLM \citep{tensorrt-llm}. Building on these foundations, Sarathi-Serve \citep{chunkedPrefill} piggybacks decode steps onto chunked prefills to keep the accelerator busy. A complementary direction disaggregates the prefill and decode phases onto separate instances: SplitWise \citep{splitwise} and DistServe \citep{pddisaggreate} show that isolating compute-bound prefill from latency-sensitive decode removes mutual interference. \system{} builds on this throughput-oriented foundation and extends it toward an explicitly SLO-aware objective.

\myline{SLO- and latency-aware scheduling.} As LLMs increasingly back Web services with diverse latency expectations, recent work shifts attention from raw throughput to Service Level Objectives. Classic inference-serving systems already framed scheduling around SLO guarantees and cost across heterogeneous hardware \citep{slo1,slo2}. In the LLM setting, Mooncake \citep{MoonCake} adopts a KVCache-centric architecture with a prediction-based early-rejection policy for overload, while QLM \citep{QueueManagement} estimates request waiting times to orchestrate queue management for TTFT adherence. Fairness-oriented schedulers further ensure equitable service across clients \citep{sheng2024fairness,shen2024fastswitch}. Closest to our setting, AdaServe \citep{Adaserve} and SLOs-Serve \citep{sloserve} adjust per-request token allocation on the fly, the latter combining chunked prefill with optional speculative decoding \citep{speculative_decoding}. We explore a new setting in which requests carry \emph{heterogeneous} TTFT and TPOT targets simultaneously, and design fine-grained guards that exploit this heterogeneity rather than treating SLOs uniformly.

\myline{Length prediction for scheduling.} Because output length is unknown a priori, a body of work predicts request characteristics to inform scheduling. Length-aware policies enable shortest-job-first style ordering, via direct length estimation \citep{s3,cheng2024enablingefficientbatchserving,bertprediction}, learning-to-rank over batches \citep{learbyrank}, or embedding-based shortest-remaining-time scheduling \citep{dontstopmenow}. Prediction has also been applied at the LMaaS layer for request management and routing \citep{jiang2025hierarchicalpredictionbasedmanagementlmaas}. \system{} uses a lightweight length predictor as one input to an analytic SLO-estimation module, turning predictions into admission-control and batch-selection decisions under heterogeneous SLOs.

\section{Conclusion} \label{sec:conclusion}

In this paper, we present \system{}, a novel SLO-oriented LLM serving system designed to maximize system goodput and SLO adherence for requests with heterogeneous SLOs. By turning SLO heterogeneity from a scheduling burden into a source of capacity, \system{} coordinates a TPOT Guard and a TTFT Guard, supported by an accurate predictor, to serve the right request at the right time. Evaluations demonstrate that \system{} significantly outperforms state-of-the-art throughput-oriented systems, improving system goodput by up to 14.4$\times$ and SLO adherence rate by up to 46.5\% across various scenarios. \system{} integrates cleanly as a scheduling layer atop existing serving stacks and complements optimizations such as prefill-decode disaggregation \citep{pddisaggreate}.

\begin{credits}
\subsubsection{\ackname}
This work was supported in part by the National Natural Science Foundation of China
(62132017, 62502430), and Zhejiang Provincial Natural Science Foundation of China (LQ26F020004).
\end{credits}

\bibliographystyle{splncs04}
\bibliography{reference}

\end{document}